\documentclass{article} % For LaTeX2e
\usepackage{iclr2027_conference,times}

\usepackage{amsmath,amsfonts,bm}

\def\eqref#1{equation~\ref{#1}}
\def\1{\bm{1}}

\DeclareMathAlphabet{\mathsfit}{\encodingdefault}{\sfdefault}{m}{sl}
\SetMathAlphabet{\mathsfit}{bold}{\encodingdefault}{\sfdefault}{bx}{n}

\usepackage{hyperref}
\usepackage{url}
\usepackage{graphicx}
\usepackage{multirow}
\usepackage{amsmath,amssymb,amsfonts,amsthm,mathtools}
\usepackage{booktabs}
\usepackage{enumitem}
\usepackage{microtype}
\usepackage{xcolor}
\usepackage{float}
\usepackage{amsmath}
\usepackage{amssymb}
\usepackage{bbm}
\usepackage{algorithm}
\usepackage{algpseudocode}
\usepackage{xcolor}
\usepackage{listings}
\usepackage{float}

\lstdefinestyle{idmprompt}{
    language={},
    basicstyle=\small\ttfamily,
    columns=fullflexible,
    keepspaces=true,
    showstringspaces=false,
    numbers=none,
    mathescape=false,
    breaklines=true,
    breakatwhitespace=true,
    breakautoindent=false,
    breakindent=0pt,
    backgroundcolor=\color{black!5},
    rulecolor=\color{black!20},
    frame=single,
    framerule=0.4pt,
    framesep=6pt,
    framexleftmargin=0pt,
    framexrightmargin=0pt,
    xleftmargin=6.4pt,
    xrightmargin=6.4pt,
    aboveskip=8pt,
    belowskip=8pt
}

\allowdisplaybreaks

\title{Predict Before You Modify:
Learning Improvement Dynamics
for Self-Evolving Agents}

\newcommand{\authorinfo}[3]{%
  \parbox[t]{\dimexpr0.5\textwidth-2\tabcolsep\relax}{%
    \normalfont\raggedright
    \textbf{#1}\\
    #2\\
    \texttt{#3}%
  }%
}

\author{%
\authorinfo
  {Zixuan Xie}
  {University of Virginia}
  {xie.zixuan@email.virginia.edu}%
\And
\authorinfo
  {Chang Xu\thanks{Corresponding authors.}}
  {Microsoft Research}
  {chanx@microsoft.com}%
\AND
\authorinfo
  {Yuhua Jiang}
  {Tsinghua University}
  {yh-jiang24@mails.tsinghua.edu.cn}%
\And
\authorinfo
  {Li Zhao\textsuperscript{*}}
  {Microsoft Research}
  {lizo@microsoft.com}%
}
\iclrfinalcopy
\begin{document}

\maketitle
\lhead{Preprint}

\begin{abstract}
Self-evolving agents improve through repeated cycles of modification and evaluation.
However, existing methods receive feedback on modifications only after costly evaluations, and aggregate scores may obscure distinctions between failures that require different repairs.
We introduce predictive self-evolution through an editable Improvement Dynamics Model (IDM).
Before evaluation, the IDM predicts task-level outcomes of a proposed modification and guides an optional revision.
After evaluation, past modifications and their observed outcomes are used to improve the IDM, allowing predictive knowledge to accumulate across generations.
The system can thus improve how it solves tasks, proposes modifications, and predicts their effects.
Building on HyperAgents, we evaluate this approach across multilingual coding, paper review, robotics reward design, and Olympiad-level mathematical grading.
Our approach outperforms HyperAgents across all four domains, with up to 46\% higher average performance over generations.
Gains also persist under matched token budgets that include both agent evolution and task evaluation, and extend to held-out tasks.
Forecast accuracy improves as the IDM evolves, and the learned predictive knowledge transfers to improve evolution in a new domain.\footnote{Code is available at
\raggedright\url{https://anonymous.4open.science/r/Predictive-self-evolution/}.}
\end{abstract}

\section{Introduction}
\label{sec:intro}

The capabilities of modern AI agents depend not only on their foundation models, but also on the programs that determine how they reason, use tools, maintain memory, and interact with their environments. Recent work on agent self-evolution seeks to improve these programs automatically by repeatedly modifying them, evaluating the resulting agents on target tasks, and using the evaluation results to guide later modifications \citep{zhang2025dgm,zhang2026hyperagents}. Yet, like Odysseus navigating without Athena, these systems rely on costly evaluations to discover the effects of their modifications. Predicting those effects could help refine proposals before evaluation.

% Existing methods make this delayed feedback useful by preserving it for later generations. The Darwin Gödel Machine (DGM) \citep{zhang2025dgm} maintains a growing archive in which evaluated coding agents can later serve as parents for new evolutionary branches. Related methods share experience across branches or give later generations access to richer evidence, including source code, evaluation scores, and execution traces \citep{weng2026group,lee2026metaharness}. HyperAgents \citep{zhang2026hyperagents} extend DGM by combining a task agent, which solves and is evaluated on the target tasks, with a meta agent, which modifies the task agent and itself, in a single editable program. 
% With access to the archive and prior evaluations, the meta agent can use experience from earlier generations to improve both task-solving behavior and the process that generates future modifications.
Past evaluations provide a basis for such predictions.
The Darwin G\"odel Machine (DGM) \citep{zhang2025dgm} maintains an
archive of evaluated coding agents that can serve as parents for new branches.
Related methods share experience across branches or expose richer
historical evidence \citep{weng2026group,lee2026metaharness}.
% HyperAgents \citep{zhang2026hyperagents} combines editable task and
% meta agents, allowing prior evaluations to improve both task-solving
% behavior and the process that generates future modifications.
% HyperAgents \citep{zhang2026hyperagents} combines a task agent that
% solves target tasks with a meta agent that uses prior evaluations
% to modify the task agent and itself.
% HyperAgents \citep{zhang2026hyperagents} combines a task agent that
% solves target tasks with a meta agent that modifies the task agent
% and itself, allowing prior evaluations to improve both task-solving
% behavior and the process that generates future modifications.
HyperAgents \citep{zhang2026hyperagents} extend DGM by combining a task agent and a meta agent in a single editable program.
The task agent solves the target tasks and is evaluated on them, while the meta agent modifies both the task agent and itself.
With access to the archive and prior evaluations, the meta agent can use experience from earlier generations to improve both task-solving behavior and the process that generates future modifications.

Recent methods use this experience to guide proposal generation or anticipate outcomes.
DREvo \citep{guo2026drevo} uses past modifications and their outcomes
to guide new proposals, while AHE \citep{lin2026ahe} records
task-level expectations and checks them against subsequent evaluations.
AgentSquare \citep{shang2025agentsquare} ranks recombined agent
candidates with an in-context performance predictor.
These approaches do not, however, use predicted task-level outcomes
as feedback for revising a draft before evaluation.
We therefore seek to turn accumulated experience into predictive feedback
that helps refine proposals before costly evaluation.

\begin{figure*}[t]
    \centering
    \includegraphics[
        width=\textwidth
    ]{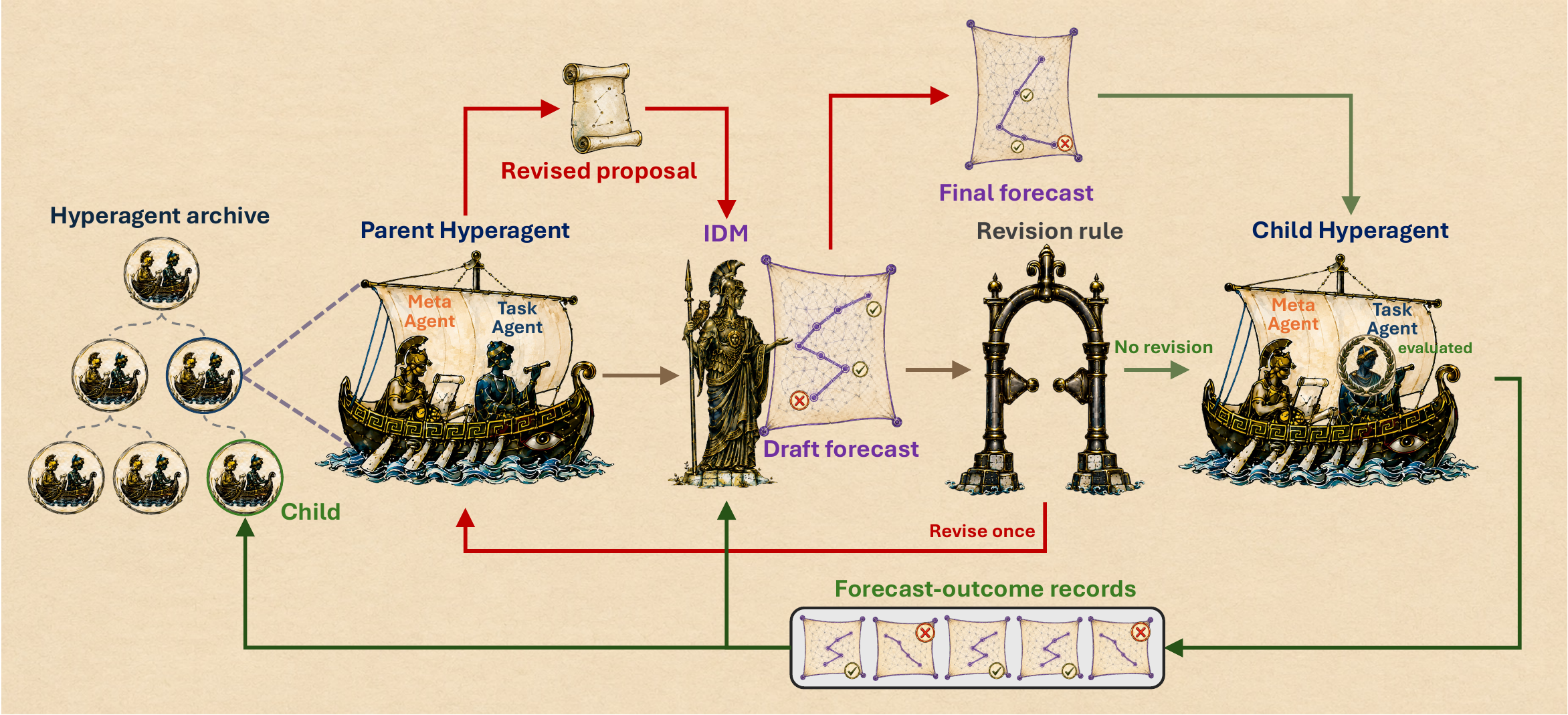}
%     \caption{
% \textbf{An Athena for self-evolution.}
% Like Athena guiding Odysseus, the shared Improvement Dynamics Model
% (IDM) helps the meta agent anticipate the consequences of a proposed
% modification before evaluation. The IDM forecasts task-level outcomes,
% and a fixed revision rule uses the forecast to decide whether the draft
% receives one additional revision. A revised proposal is forecast again
% before the child is constructed. Only the task agent in the resulting
% child is evaluated. The forecast for the final modification is
% recorded with the observed outcomes, providing evidence for improving
% the shared IDM in later generations.
% }
    \caption{
\textbf{An Athena for self-evolution.}
Like Athena guiding Odysseus, the IDM anticipates task-level outcomes
to guide an optional revision before evaluation.
Only the resulting child's task agent is evaluated.
Final forecasts and observed outcomes provide evidence for improving
the IDM across generations.
}
    \label{fig:overview}
\end{figure*}

Using predictions to guide revision requires anticipating how a
modification will affect individual tasks, not just whether it will
improve the overall score.
% Building on HyperAgents \citep{zhang2026hyperagents}, we introduce
% predictive self-evolution with a shared, editable Improvement
% Dynamics Model (IDM), making the prediction procedure itself
% an object of evolution.
Building on HyperAgents \citep{zhang2026hyperagents}, we introduce
predictive self-evolution with an Improvement Dynamics Model (IDM),
an editable program that invokes a frozen LLM to forecast task-level
outcomes from the current agent, a proposed code change, and historical
experience.
These forecasts preserve distinctions relevant to revision:
an incorrect attempt and a missing output, for example, may receive
the same failure score but call for different repairs.
The predicted aggregate score determines whether to request one
additional revision from the meta agent, while the task-level
forecasts guide that revision.
The resulting descendant's task agent is then evaluated.
% The meta agent periodically modifies the shared IDM using past
% modifications and their predicted and observed outcomes from across
% generations and branches.
To improve forecasts as the agents themselves evolve,
the meta agent periodically edits
the IDM's forecasting code and instructions using past modifications
and discrepancies between their predicted and observed outcomes
from across generations and branches.
Candidate updates are adopted only if they improve prediction
quality on subsequently collected outcomes.
Figure~\ref{fig:overview} illustrates this process.

% Our contributions are as follows.
Following HyperAgents \citep{zhang2026hyperagents}, we evaluate
predictive self-evolution across four domains: multilingual coding
\citep{gauthier2024polyglot}, paper review \citep{zhao2026apres},
robotics reward design \citep{genesis2024}, and Olympiad-level
mathematical grading \citep{luong2025towards}.
Our contributions are as follows.
\begin{itemize}
    \item \textbf{Predictive self-evolution.}
    We introduce an editable IDM that forecasts
    task-level outcomes of proposed modifications
    and guides selective revision before evaluation.
    Its forecasting program evolves from experience
    alongside the task and meta agents.

    \item \textbf{Effective use of evolution budgets.}
    Allocating part of the budget to predictive feedback
    improves evolution across four domains under matched
    token budgets covering evolution and task evaluation.
    Coding ablations support its value beyond an
    additional revision alone.

    \item \textbf{Transferable predictive knowledge.}
    Evolution produces predictive knowledge reusable
    beyond the agents that generated it.
    IDM forecasts improve with experience, and transferring
    only the evolved IDM improves mathematical-grading
    evolution over a freshly initialized IDM, with
    identical initial task and meta agents.
\end{itemize}

\section{Related Work}
\label{sec:related}

\paragraph{Automatic agent design and self-evolution.}
Automatic agent design uses task feedback to search for effective programs and workflows built around foundation models.
Automated Design of Agentic Systems (ADAS) \citep{hu2025adas} uses a meta agent to discover agent programs in code, while AFlow \citep{zhang2025aflow} searches over executable workflows using Monte Carlo tree search.
Reusing earlier discoveries is central to these searches.
The Darwin G\"odel Machine \citep{zhang2025dgm} maintains an archive of evaluated coding agents that can serve as starting points for further self-modification, while Group-Evolving Agents \citep{weng2026group} organize evolution around groups that share experience across agents.
HyperAgents \citep{zhang2026hyperagents} extend evolution to the meta agent responsible for generating modifications, allowing the improvement procedure itself to change alongside the task agent.
% Building on this framework, we introduce a shared IDM to guide modifications before evaluation.
We extend HyperAgents with an editable IDM that evolves alongside the task and meta agents.

\paragraph{Learning from feedback and experience.}
Self-Refine \citep{madaan2023selfrefine} uses self-generated
feedback to revise the current output, while Reflexion
\citep{shinn2023reflexion} retains verbal reflections to guide
later trials.
Other approaches organize experience for repeated reuse:
Agentic Context Engineering (ACE) \citep{zhang2026ace}
maintains evolving contextual playbooks, and ReasoningBank
\citep{ouyang2026reasoningbank} distills reusable reasoning
strategies from successful and failed experiences.
Knowledge-Centric Self-Improvement (KSI) \citep{wang2026ksi}
accumulates insights in a shared knowledge base that generic
agents use across tasks.
Beyond supplying context for task execution, feedback can
also guide changes to the agent's prompts and programs.
GEPA \citep{agrawal2026gepa} reflects on execution trajectories
to propose and test prompt updates and combine complementary
improvements.
For program modification, Meta-Harness
\citep{lee2026metaharness} gives the optimizer access to
previous candidates' source code, scores, and execution traces.
HarnessFix \citep{chen2026harnessfix} and Self-Harness
\citep{zhang2026selfharness} use execution traces to diagnose
failures and validate targeted harness repairs, whereas
DREvo \citep{guo2026drevo} reassesses historical evidence
against the current program before distilling it into
search directions.

\paragraph{Predictive models for agent improvement.}
Predictive models have long been used to guide search and
reduce evaluation costs, including in surrogate-assisted
evolutionary optimization \citep{jin2011surrogate} and
predictor-based neural architecture search \citep{luo2018neural}.
At the program level, Neural Change Prediction
\citep{plein2026neuralchange} learns from synthetic code
mutations and their observed effects to predict changes
in program outputs.
At the task-execution level, WorldEvolver
\citep{zhang2026worldevolver} forecasts environment
transitions to guide actions and updates predictive memory
from discrepancies between forecasts and observations.
For agent optimization, predictions can support candidate
selection or make the expected effects of edits explicit.
AgentSquare \citep{shang2025agentsquare} uses an in-context
performance predictor to rank candidates formed by
recombining existing agent modules, reducing the need
for direct evaluation.
AHE \citep{lin2026ahe} records expected fixes and regressions
for each edit and checks these predictions against
subsequent task-level outcomes.
Our IDM instead uses forecasts of task-level evaluation
outcomes as feedback for revising a proposed agent-program
modification before evaluation.
The meta agent also edits the IDM's forecasting code
and instructions using past modifications and prediction errors.

\section{Predictive Self-Evolution}
\label{sec:problem-formulation}
% \subsection{Archive-Based Self-Evolution}
% \label{sec:archive-based}
% \paragraph{Agents and task evaluation.}

% Following Hyperagents \citep{zhang2026hyperagents}, we define an \emph{agent} as any computable program.
% A \emph{task agent} produces outputs for target tasks, whereas a \emph{meta agent} generates new agents by modifying existing agent programs.
% A \emph{Hyperagent} combines a task agent and a meta agent in a single editable program.
% Its meta agent may modify any editable part of this program, including the task agent and itself, to produce a new Hyperagent.

% For a fixed target domain $d$, let $\mathcal T_d$ denote its finite set of evaluation tasks.
% Evaluating a Hyperagent $h$ runs its task agent on each task $\tau\in\mathcal T_d$.
% The evaluation score is the fraction of tasks on which its task agent succeeds:
% \begin{equation}
% y(h)
% \doteq
% \frac{1}{|\mathcal T_d|}
% \sum_{\tau\in\mathcal T_d}
% \mathbbm{1}
% \left\{\text{the task agent in $h$ succeeds on $\tau$}\right\}.
% \label{eq:evaluation-score}
% \end{equation}

\subsection{Archive-Based Self-Evolution}
\label{sec:archive-based}
\paragraph{Agents and task evaluation.}

% Following HyperAgents \citep{zhang2026hyperagents}, we define an
% \emph{agent} as any computable program. 
% A \emph{task agent} receives target tasks and produces task outputs,
% whereas a \emph{meta agent} generates new agents by modifying existing ones.
% A \emph{task agent} solves target tasks, whereas a \emph{meta agent} generates new agents by modifying existing agents and generates new ones.
% A \emph{Hyperagent} integrates a task agent and a meta agent into a single editable program. 
% Its meta agent may modify any editable part of the program, including the implementations of the task agent and the meta agent itself, thereby producing a new Hyperagent.
Following HyperAgents \citep{zhang2026hyperagents}, an \emph{agent}
is any computable program.
A \emph{Hyperagent} combines a \emph{task agent} that solves
target tasks with a \emph{meta agent} that can modify both
the task agent and itself.
For a fixed target domain $d$, let $\mathcal T_d$ denote its finite set of evaluation tasks.
Evaluating a Hyperagent $h$ runs its task agent under the domain-specific evaluation protocols of HyperAgents \citep{zhang2026hyperagents}.
Let $s_\tau(h)$ denote the resulting score on task $\tau\in\mathcal T_d$, with higher scores indicating better performance.
We define the aggregate evaluation score as
\begin{equation}
y(h)
\doteq
\frac{1}{|\mathcal T_d|}
\sum_{\tau\in\mathcal T_d}
s_\tau(h).
\label{eq:evaluation-score}
\end{equation}
Task scores may be binary or continuous and are averaged over repeated trials where required by the evaluation protocol.
Domain-specific evaluation protocols and score aggregation
are detailed in Appendix~\ref{app:domains}.

\paragraph{Evolution process.}

% A run begins with an evaluated Hyperagent $h_0$.
% At generation $t$, let $\mathcal A_t$ denote the archive of evaluated Hyperagents.
% The archive forms a rooted tree with $h_0$ as its root.
% Each node stores a Hyperagent and its evaluation score, and each edge records a parent--child relation.
% At each generation, a parent Hyperagent $h_{p_t}$ is selected using the fixed parent-selection rule of HyperAgents \citep{zhang2026hyperagents}.
% Its meta agent uses the archive and prior evaluation results to propose a code diff $\delta_t$.
% Applying this diff to a copy of the parent produces a child Hyperagent $h_{t+1}$, which is evaluated and added to the archive.
% Earlier hyperagents remain available as parents, allowing evolution to explore multiple branches and build on experience from previous generations.
% Our method introduces predictive feedback into this loop, allowing the meta agent to revise a proposed diff before the resulting child is evaluated.
At generation $t$, the archive $\mathcal A_t$ forms a tree rooted
at the evaluated initial Hyperagent $h_0$, storing agents, scores,
and parent--child relations.
A parent $h_{p_t}$ is sampled using the fixed HyperAgents selection
rule \citep{zhang2026hyperagents}.
Its meta agent uses the archive and prior evaluations
to propose a code diff $\delta_t$.
Applying this diff to a copy of the parent produces
a child $h_{t+1}$, which is evaluated and added to
the archive.
Earlier agents remain available as parents.
Our method adds predictive feedback to guide an optional
revision of the diff before child evaluation.

\subsection{Improvement Dynamics Model}
\label{sec:idm}

\begin{figure*}[t]
    \centering
    \includegraphics[
        width=\textwidth
    ]{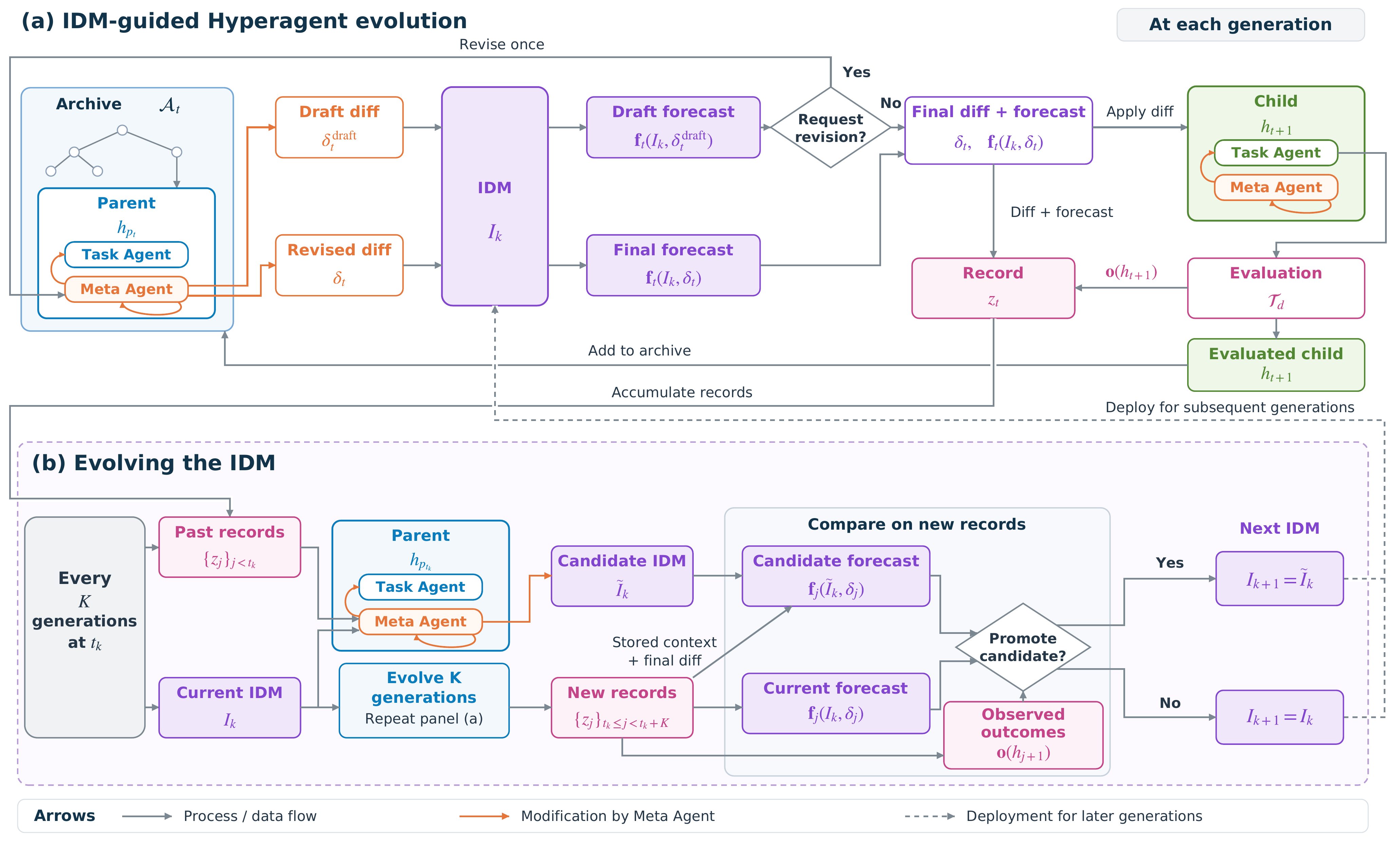}
%     \caption{
%     \textbf{Pipeline of predictive self-evolution.}
%     At each generation, the meta agent in the selected parent proposes a draft diff, and the shared IDM forecasts the task-level outcomes of the resulting child. 
%     A fixed revision rule either accepts the draft or returns the forecast for one revision. 
%     If revised, the final diff is forecast again before evaluation. 
%     The final forecast and observed outcomes form a record, and the evaluated child enters the Hyperagent archive.
%     Periodically, the meta agent uses accumulated records to produce a candidate IDM, which replaces the current IDM only after validation on fresh records.
% }
    \caption{
    \textbf{Pipeline of predictive self-evolution.}
    Upper: the IDM forecasts a draft and guides at most one revision;
    revised diffs are forecast again before evaluation.
    The final forecast and observed outcomes form a record, and the
    child enters the archive.
    Lower: accumulated records support candidate IDM updates, which
    replace the current IDM only after validation on subsequently
    collected records.
}
    \label{fig:predictive-self-evolution}
\end{figure*}

% We introduce a shared, editable Improvement Dynamics Model (IDM).
% The IDM is a program that invokes a frozen LLM, with editable code
% specifying its forecasting procedure and instructions.
% At each generation, it predicts the task-level outcomes of a
% proposed modification to guide revision before evaluation.
% A single IDM serves all archive branches and is periodically updated
% using past modifications and their observed outcomes.
% Figure~\ref{fig:predictive-self-evolution} illustrates these processes.
The IDM $I$ is an editable program that invokes a frozen LLM
to forecast task-level outcomes before evaluation.
A single IDM serves all archive branches within a run.
Its forecasting code and instructions evolve from past
modifications and observed outcomes
(Figure~\ref{fig:predictive-self-evolution}).
Let $\mathbf o(h)$ denote the task-level outcomes recorded when
evaluating a Hyperagent $h$.
At generation $t$, the IDM receives the archive $\mathcal A_t$,
the selected parent $h_{p_t}$ and its observed outcomes
$\mathbf o(h_{p_t})$, and a proposed code diff $\delta$.
Writing $h_{p_t}^{\delta}$ for the child that would result from
applying this diff, we denote the IDM's forecast by
\begin{equation}
    \mathbf f_t(I,\delta)
    \doteq
    I\!\left(
        \mathcal A_t,
        h_{p_t},
        \mathbf o(h_{p_t}),
        \delta
    \right).
    \label{eq:idm-prediction}
\end{equation}

The forecast $\mathbf f_t(I,\delta)$ contains task-level
outcome probabilities and, where applicable, continuous
performance estimates.
These forecasts distinguish outcomes that may receive
the same task score but call for different repairs,
such as an incorrect solution and a missing output.
Domain-specific prediction targets and their conversion to task scores are detailed in Appendix~\ref{app:idm-implementation}.

Let $\widehat s_{t,\tau}(I,\delta)$ denote the expected task score
derived from the forecast for task $\tau$.
Following the aggregation used in task evaluation, the predicted
score of the proposed child is
\begin{equation}
    \widehat y_t(I,h_{p_t}^{\delta})
    \doteq
    \frac{1}{|\mathcal T_d|}
    \sum_{\tau\in\mathcal T_d}
    \widehat s_{t,\tau}(I,\delta).
    \label{eq:idm-predicted-score}
\end{equation}
The predicted score determines whether to request one additional
revision, while the task-level forecast guides that revision.

\subsection{IDM-Guided Hyperagent Evolution}
\label{sec:idm-guided-evolution}

As illustrated in the upper part of
Figure~\ref{fig:predictive-self-evolution}, each generation $t$
begins with the meta agent in the selected parent $h_{p_t}$
proposing a draft diff $\delta_t^{\mathrm{draft}}$.
The current IDM $I$ remains fixed during this generation
and produces the draft forecast
$\mathbf f_t(I,\delta_t^{\mathrm{draft}})$.
We correct the predicted score by subtracting an online
estimate $b$ of the current IDM's prediction bias.
Calibration details are provided in
Appendix~\ref{app:idm-update-validation}.
A fixed revision rule shared across all branches
requests one additional revision when
\begin{equation}
    \widehat y_t
    \left(
        I,h_{p_t}^{\delta_t^{\mathrm{draft}}}
    \right)
    -
    b
    <
    y(h_{p_t}).
    \label{eq:revision-gate}
\end{equation}
The revision rule allocates an additional revision to drafts
whose bias-corrected predicted scores fall below the
parent's observed score.
% For coding, both scores use the same ten preliminary tasks
% (Appendix~\ref{app:domains}).
When revision is requested, the meta agent receives the draft
forecast and produces a revised diff $\delta_t$.
Otherwise, $\delta_t=\delta_t^{\mathrm{draft}}$.
Following a revision, the IDM forecasts the final diff once more
so that the recorded prediction corresponds to the modification
actually evaluated.
To limit computation, we allow at most one revision
before obtaining fresh feedback from child evaluation.

The final diff and its forecast $\mathbf f_t(I,\delta_t)$ are
recorded before evaluating the task agent in the resulting child
$h_{t+1}=h_{p_t}^{\delta_t}$.
This evaluation proceeds regardless of the final predicted score.
The evaluated child is added to the archive, and the completed
generation yields the record
\begin{equation}
    z_t
    \doteq
    \left(
        \mathcal A_t,
        h_{p_t},
        \mathbf o(h_{p_t}),
        \delta_t,
        \mathbf f_t(I,\delta_t),
        \mathbf o(h_{t+1})
    \right).
    \label{eq:idm-record}
\end{equation}
The record preserves the inputs available to the IDM before
evaluation and pairs the final forecast with the observed child
outcomes for subsequent IDM updates.

\subsection{Evolving the IDM}
\label{sec:evolving-idm}

Task performance guides the evolution of task and meta agents,
but does not directly measure the IDM's prediction accuracy.
We therefore update the IDM separately from the task and meta
agents and share it across archive branches, allowing records
from different branches to improve the same forecasting program.
As illustrated in the lower part of
Figure~\ref{fig:predictive-self-evolution}, every $K \geq 2$
generations, the meta agent in the selected Hyperagent uses
the accumulated records to edit the current IDM's forecasting
code and instructions, producing a candidate.
We validate the candidate on the next $K$ records, collected
while the current IDM remains deployed.
Once the block is complete, the candidate makes its forecasts
using the recorded pre-evaluation inputs, without access to
later archive updates or the corresponding child outcomes.

To replace the current IDM, a candidate must reduce prediction
loss in the task forecasts used to guide revision.
We measure this loss using Brier loss
\citep{brier1950verification} for probabilities and squared
error for continuous values, with domain-specific definitions
given in Appendix~\ref{app:idm-implementation}.
For the score that determines whether to revise, we also require
that centered mean squared error (CMSE)
\citep{taylor2001summarizing} does not increase.
CMSE measures score prediction error after removing each IDM's
mean bias on the validation block.
If both conditions hold, the candidate replaces the current IDM;
otherwise, the current IDM remains deployed.
Upon replacement, the candidate's mean bias on this block
initializes the online calibration offset $b$ described in
Section~\ref{sec:idm-guided-evolution}.

\section{Experiments}
\label{sec:experiments}

% We evaluate whether predictive self-evolution improves agent performance under generation and token budgets and whether learned predictive knowledge transfers to a new domain.
We evaluate whether predictive self-evolution improves agent performance under generation and token budgets.
% We then examine the contributions of its main components and how forecast accuracy changes during evolution.
% Coding ablations examine component contributions; we then study forecast accuracy during evolution.
% We then examine the contributions of individual components through coding ablations and analyze how forecast accuracy changes during evolution.
% We then examine how forecasts improve during evolution, whether learned predictive knowledge transfers to a new domain, and how individual components contribute to performance.
We then examine forecast accuracy during evolution,
cross-domain transfer of predictive knowledge, and the
contributions of individual components through coding ablations.

\subsection{Experimental Setup}
\label{sec:experimental-setup}
\paragraph{Initialization.}
We use the initial Hyperagent from HyperAgents
\citep{zhang2026hyperagents}, adding instructions that distinguish
Hyperagent modifications from IDM updates and specify
the feedback available in each phase.
The initial IDM is instructed to predict task-level
outcomes from proposed modifications and available evidence.
Implementation details and instruction templates are
provided in Appendix~\ref{app:experimental-details}.

\paragraph{Domains and evaluation protocol.}
Following HyperAgents \citep{zhang2026hyperagents}, we evaluate predictive self-evolution across four domains:
Polyglot coding \citep{gauthier2024polyglot},
paper review \citep{zhao2026apres},
robotics reward design in Genesis \citep{genesis2024},
and Olympiad-level mathematical grading using IMO-GradingBench \citep{luong2025towards}.
Polyglot is evaluated independently, paper review and robotics are optimized jointly, and mathematical grading is the target setting for IDM transfer.
Each method is evaluated over five complete runs per setting,
following HyperAgents \citep{zhang2026hyperagents}.
Each run lasts 40 generations.
Within each setting, methods share the same initial Hyperagent, underlying task-agent and meta-agent LLMs, task splits, evaluation protocol, and parent-selection rule.
Domain-specific evaluation protocols are detailed in Appendix~\ref{app:experimental-details}, and IDM prediction formats and losses in Appendix~\ref{app:idm-implementation}.

\paragraph{Baselines and variants.}
We refer to our method as \emph{Selective IDM} in the experiments.
HyperAgents serves as the common baseline across the domains.
It evaluates the modified agent without requesting an additional revision from the meta agent.
On Polyglot, Reflective HyperAgents revises every draft once without IDM feedback.
We also evaluate two variants of our method on this benchmark.
Frozen Selective IDM retains selective revision and online bias calibration but fixes the IDM to its initial implementation.
IDM Reflection uses feedback from the evolving IDM to revise every draft once.
For mathematical grading, Selective IDM starts from the initial IDM, while Transferred IDM starts from an IDM evolved jointly on paper review and robotics.
Both conditions use the same evolution procedure and initial task and meta agents.

\paragraph{Metrics and budgets.}
Task performance is measured by the evaluation score defined for each domain.
The joint score is the equally weighted mean of paper-review
validation accuracy and robotics training fitness.
To assess evolution efficiency, we use the normalized area under the budget curve (AUBC), which captures the benefit of discovering strong agents earlier.
Let $F_t$ denote the highest score among evaluated descendants available after generation $t$.
For a generation budget $B$, we define
\begin{equation}
    \operatorname{AUBC}(B)
    \doteq
    \frac{1}{B}\sum_{t=0}^{B-1}F_t,
    \qquad F_0=0.
    \label{eq:aubc}
\end{equation}
For matched-token comparisons, we compute AUBC at a common budget
equal to the reference baseline's median full-run token cost.
The reference is HyperAgents for the main comparisons and
Reflective HyperAgents for the coding ablations.
Token costs cover agent evolution and task evaluation.
Token accounting and budget matching are detailed in Appendix~\ref{app:evaluation-cost-analysis}.
Performance metrics are computed within each run before aggregation, and we report medians over five complete runs.
Individual-run results are provided in Appendix~\ref{app:run-level-results}.

\subsection{Improving Agent Evolution}
\label{sec:task-performance}

\begin{figure}[t]
    \centering
    \includegraphics[width=\textwidth]{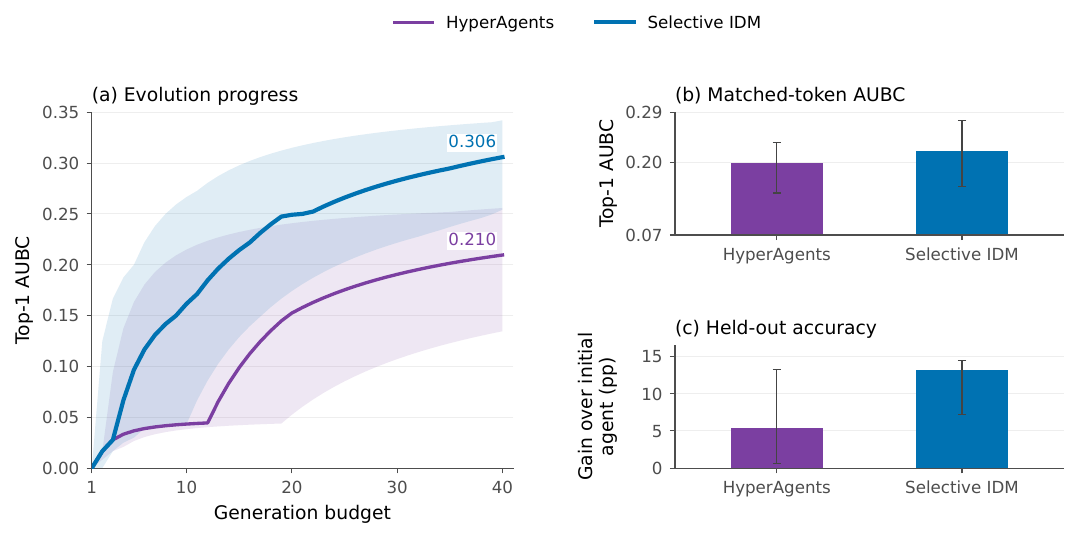}
    \caption{
    \textbf{More effective coding evolution with gains
    on held-out tasks.}
    (a) AUBC as the generation budget increases.
    (b) AUBC under matched token budgets, including
    tokens used for both agent evolution and task
    evaluation.
    % The vertical axis in (b) is truncated.
    (c) Improvement in resolve rate over the common
    initial agent on 166 held-out coding tasks,
    measured in percentage points (pp).
    The initial agent's resolve rate is 16.27\%.
    Curves and bars show medians over five runs.
    Shading and error bars indicate 95\% run-level
    bootstrap confidence intervals, computed pointwise in (a).
    }
    \label{fig:coding-main}
\end{figure}

\paragraph{Coding.}
Using predicted outcomes to guide revision before evaluation helps Selective IDM discover stronger agents earlier.
On Polyglot, it achieves a median AUBC of 0.306 over 40 generations, compared with 0.210 for HyperAgents
(Figure~\ref{fig:coding-main}(a)).
Under matched token budgets, median AUBC remains 10.9\% higher than that of HyperAgents
(Figure~\ref{fig:coding-main}(b)).
This gain supports using part of the token budget for predictive feedback before evaluating modified agents.

On 166 held-out coding tasks, agents selected from Selective IDM runs achieve a median gain of 13.25 percentage points over the common initial agent, compared with 5.42 points for HyperAgents
(Figure~\ref{fig:coding-main}(c)).
These results show that the gains from predictive self-evolution extend to tasks not used for modification or agent selection.

\begin{figure}[t]
    \centering
    \includegraphics[width=\linewidth]{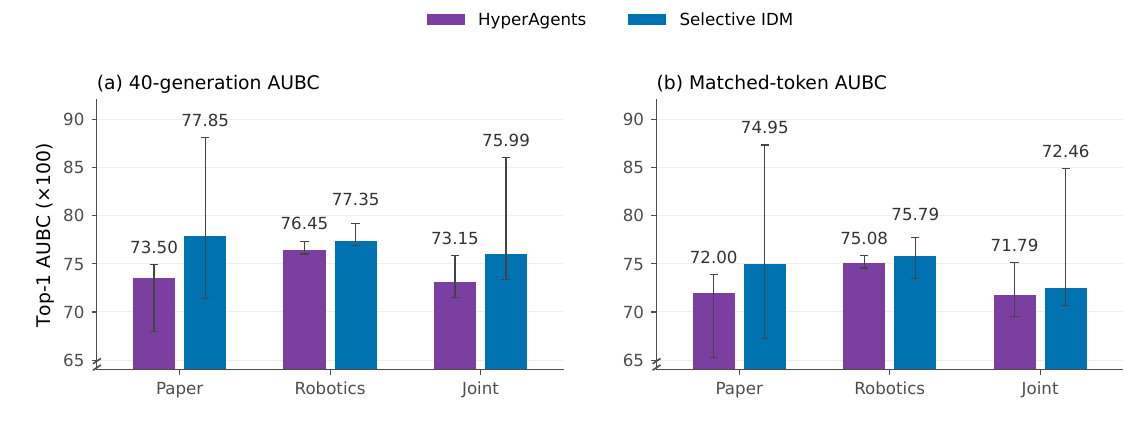}
    \caption{
    \textbf{Predictive guidance improves joint evolution
    across different task objectives.}
    AUBC under (a) a budget of 40 generations and
    (b) matched token budgets.
    Paper and Robotics track the best evaluated
    descendant for each domain.
    Joint tracks the highest joint score attained
    by a single evaluated descendant.
    Bars show medians over five runs, and error bars
    indicate 95\% run-level bootstrap confidence intervals.
    AUBC values are multiplied by 100.
    The vertical axes are truncated.
    }
    \label{fig:joint-performance}
\end{figure}

\paragraph{Paper review and robotics.}
% Joint evolution tests whether a shared IDM can guide improvement across tasks with different evaluation signals.
Joint evolution tests whether an IDM can guide improvement across tasks with different evaluation signals.
We jointly optimize paper review and robotics reward design, with equal weight on paper-review validation accuracy and robotics fitness.
The predicted joint score determines whether to request a revision, while task-level forecasts from both domains guide the revision.
Over 40 generations, Selective IDM achieves higher median AUBC than HyperAgents for both domains and the joint objective
(Figure~\ref{fig:joint-performance}(a)).
On paper review, median AUBC improves by 5.9\% over HyperAgents.
Median joint AUBC also improves by 3.9\%.
The advantage persists across all three measures under matched token budgets
(Figure~\ref{fig:joint-performance}(b)).
% Together with the coding results, these findings support using a shared predictive model to guide evolution across different task objectives.
Together with the coding results, these findings support using an IDM to guide evolution across different task objectives.

\subsection{Improving Forecasts during Evolution}
\label{sec:idm-improvement}

\begin{figure}[!htbp]
    \centering
    \includegraphics[width=\linewidth]{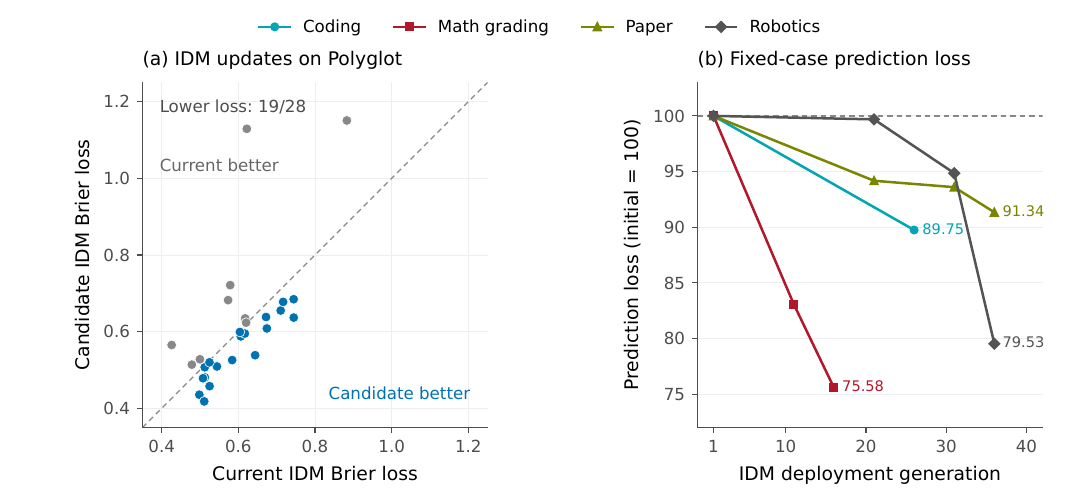}
\caption{
\textbf{IDM predictions improve during evolution.}
(a) Candidate and current IDMs are compared on subsequent
validation records in 28 comparisons across five Polyglot runs,
including rejected candidates.
Blue points indicate lower candidate Brier loss.
(b) Successive deployed IDMs are evaluated on fixed modifications
from generations 1--5 of selected runs, with initial loss
normalized to 100.
Markers indicate each version's first deployment.
Mathematical grading uses Selective IDM.
}
    \label{fig:idm-improvement}
\end{figure}

We first examine whether IDM updates improve predictions for
subsequent modifications.
On Polyglot, we compare candidate and current IDMs on the same
modification records collected after candidate construction.
Across five runs, candidates achieve lower Brier loss in
19 of 28 comparisons, including both adopted and rejected
% candidates (Figure~\ref{fig:idm-improvement}(a)).
% candidates (Figure~\ref{fig:idm-improvement}(a); run-level loss
% changes in Table~\ref{tab:idm-update-magnitudes}).
candidates (Figure~\ref{fig:idm-improvement}(a)).
Editing the IDM using past experience
can improve its predictions beyond the modifications used
to construct the update.

% To examine how prediction quality changes across deployed
% versions, we additionally evaluate successive IDMs on fixed
% historical modifications from generations 1--5 of selected runs.
We next compare successive deployed IDMs on a fixed set of
historical modifications from generations 1--5 of selected runs.
% historical modifications from generations 1--5 of selected runs
% (Appendix~\ref{app:forecast-evolution-evidence}).
The inputs, observed outcomes, and scoring rules remain
the same across versions.
In these comparisons across four domains, the final deployed
% version has lower loss than the initial IDM, with reductions
version has lower task-level prediction loss than the initial IDM, with reductions
of 8.7\%--24.4\% (Figure~\ref{fig:idm-improvement}(b)).
These results support accumulating reusable predictive
knowledge through IDM evolution.
Appendices~\ref{app:modification-information}
and~\ref{app:evolution-case-studies} examine modification
information, forecast-guided revision, and IDM evolution.

\subsection{Transferring Predictive Knowledge across Domains}
\label{sec:idm-improvement-transfer}

To test whether predictive experience remains useful in a new domain, we transfer an IDM evolved jointly on paper review and robotics to mathematical grading.
% We compare HyperAgents, Selective IDM, and Transferred IDM, a variant of Selective IDM initialized with the evolved IDM.
We compare HyperAgents with two conditions: Selective IDM
starts from the initial IDM, whereas Transferred IDM uses
the same evolved IDM in all five transfer runs.
Only the IDM is transferred, while all three conditions use the same initial task and meta agent implementations.
Both conditions continue updating the IDM.

\begin{figure}[t]
    \centering
    \includegraphics[width=\linewidth]{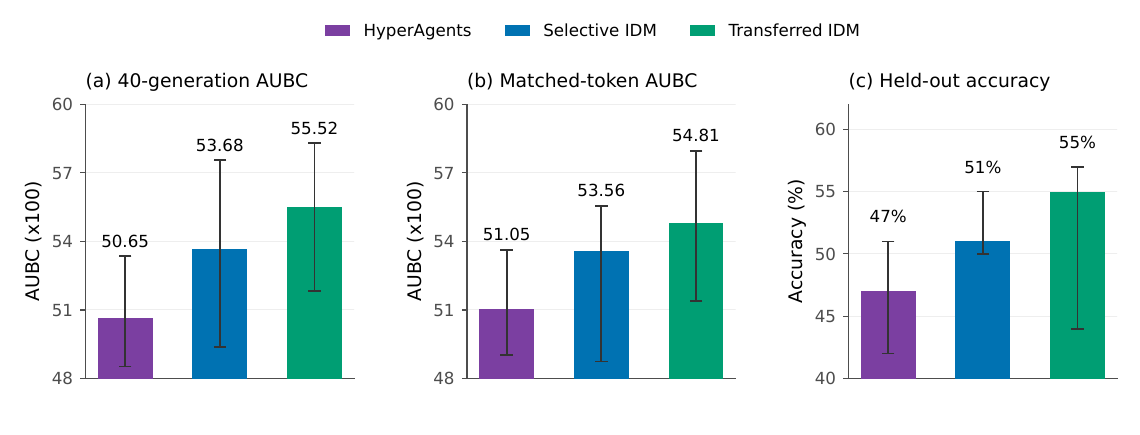}
    \caption{
    \textbf{Predictive experience transfers to
    mathematical grading.}
    AUBC under (a) 40 generations and
    (b) a matched token budget for mathematical grading.
    (c) Accuracy on 100 held-out tasks for agents
    selected by validation performance.
    Bars show medians over five runs, with
    95\% run-level bootstrap confidence intervals.
    AUBC values are multiplied by 100.
    The vertical axes are truncated.
    }
    \label{fig:math-transfer}
\end{figure}

Selective IDM achieves higher median AUBC than HyperAgents under both generation and token budgets.
Transferred IDM improves further, achieving 3.4\% higher median AUBC than Selective IDM under the generation budget and 2.3\% higher under the matched token budget for mathematical grading
(Figure~\ref{fig:math-transfer}(a,b)).
The benefit extends to held-out tasks, where Transferred IDM increases median accuracy by 4 percentage points over Selective IDM and 8 points over HyperAgents (Figure~\ref{fig:math-transfer}(c)).
These results support using the IDM to carry predictive experience across domains and guide the evolution of newly initialized agents.

\subsection{Ablations and Revision Strategies}
\label{sec:ablations}

\begin{figure}[t]
    \centering
    \includegraphics[width=0.95\linewidth]{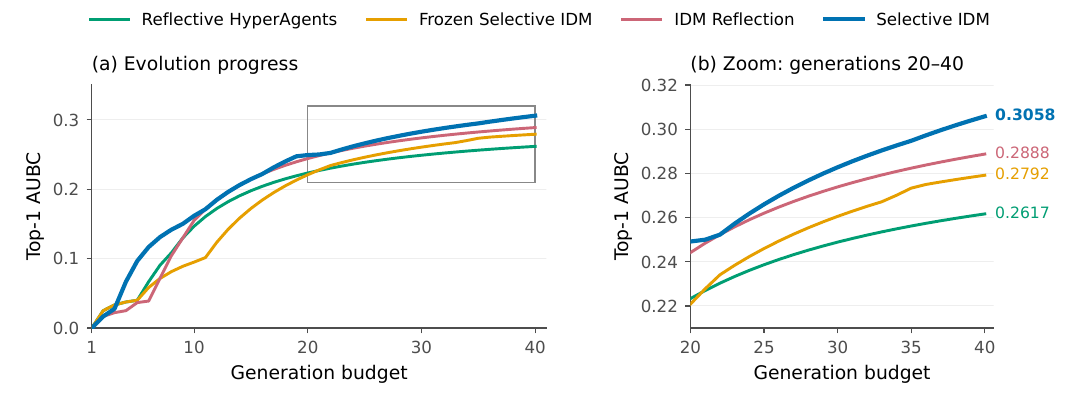}
    \caption{
    \textbf{Effects of predictive feedback, IDM updates,
    and revision strategy.}
    Curves show median AUBC over five runs on Polyglot
    as the generation budget increases.
    The left panel shows the complete trajectory;
    the right panel enlarges the outlined region
    for budgets of 20--40 generations.
    }
    \label{fig:coding-ablations}
\end{figure}

On Polyglot, we compare IDM Reflection with Reflective HyperAgents
to assess predictive feedback beyond revision alone.
Both methods revise every draft once.
IDM Reflection achieves a higher median AUBC at 40 generations, supporting the value of predictive feedback
(Figure~\ref{fig:coding-ablations}).
% Across all 40 generations, IDM Reflection also achieves a higher average descendant resolve rate on the preliminary tasks, with medians across runs of 36.50\% versus 29.25\%.
% We also average resolve rates on the same ten preliminary tasks over all 40 descendants within each run.
% The median of these averages is 36.50\% for IDM Reflection and 29.25\% for Reflective HyperAgents.
This benefit extends beyond the best agents to the quality of generated descendants (Appendix~\ref{app:descendant-quality}).
We next compare Selective IDM with Frozen Selective IDM, which
retains selective revision and online bias calibration but keeps
the forecasting program fixed.
Updating the IDM yields a higher median AUBC.

% Selective revision achieves a higher median AUBC with fewer revision passes.
Both Selective IDM and IDM Reflection use an evolving IDM,
but Selective IDM gains an early AUBC advantage.
At 40 generations, it achieves the highest median AUBC among the
four revision configurations while requesting a median of 27
additional revision passes per run, compared with 40 for IDM Reflection.
Its median AUBC remains highest under a matched token budget
(Appendix~\ref{app:evaluation-cost-analysis}).

\section{Conclusion}
\label{sec:conclusion}

We introduced predictive self-evolution with an editable
IDM that learns from past modifications and their outcomes
to guide revision before evaluation.
By making the forecasting program itself an object of evolution,
the system can accumulate and reuse predictive knowledge
as the agents continue to change.
Experiments across four domains show more effective
evolution under matched budgets and support the transfer
of predictive knowledge to a new domain.
Forecast accuracy also improves with experience.
Extending the IDM toward a world model of agent evolution
could help identify modification sequences whose early steps
offer little immediate gain but enable greater improvements
in later generations.

\subsection*{AI use statement}

% (This section is \textbf{required} and does not count toward the page limit.)
%
% In this work, we used generative AI tools for [tasks with required disclosure].
% We have not used generative AI tools for [other tasks with required disclosure],
% and [the rest of the required disclosure tasks] are not applicable to this work.
% Additionally, we used generative AI tools for [tasks with recommended
% disclosure]. We have reviewed all AI-assisted work. [Elaborate. For example, “we
% checked LLM-generated research ideas for potential plagiarism through a manual
% literature survey”, “LLM-generated code was verified and tested for correctness
% by 2 authors”, etc.]. We take responsibility for the final content of this work,
% including text, claims or artifacts produced with the aid of generative AI.
%
% See the ICLR 2027 AI Policy for Authors for more details. This statement should
% not be more than 1 page.
% We used generative AI tools to assist with implementation, analysis
% and interpretation of results, figure preparation, and manuscript
% drafting and editing.
% The authors reviewed the AI-assisted material and take responsibility
% for the final content.
In this work, we used generative AI tools for method implementation
and analysis and interpretation of results.
We have not used generative AI tools to develop research ideas or
design the methodology or experiments, and synthetic dataset
generation and mathematical proofs are not applicable to this work.
Additionally, we used generative AI tools for figure preparation
and manuscript drafting and editing.
We have reviewed all AI-assisted work and take responsibility for
the final content, including AI-assisted text, code, and analyses.

\subsection*{Ethics statement}

% (This section is \textbf{recommended} and does not count toward the page limit.)
%
% If authors feel that their paper submission raises questions regarding the Code
% of Ethics, they are encouraged to include a paragraph of Ethics Statement (at
% the end of the main text before references) to address potential concerns where
% appropriate. Topics include, but are not limited to, studies that involve human
% subjects, practices to data set releases, potentially harmful insights,
% methodologies and applications, potential conflicts of interest and sponsorship,
% discrimination/bias/fairness concerns, privacy and security issues, legal
% compliance, and research integrity issues (e.g., IRB, documentation, research
% ethics). This statement should not be more than 1 page.
We evaluate agent self-evolution on existing benchmarks, with robotics
experiments conducted in simulation.
Paper-review and mathematical-grading results concern benchmark
performance, not suitability for replacing human judgment.
Deployment beyond these settings would require safeguards for generated
code and human oversight of consequential decisions.

\subsection*{Reproducibility statement}

% (This section is \textbf{recommended} and does not count toward the page limit.)
%
% It is important that the work published in ICLR is reproducible. Authors are
% strongly encouraged to include a paragraph-long Reproducibility Statement at the
% end of the main text (before references) to discuss the efforts that have been
% made to ensure reproducibility. This paragraph should not itself describe
% details needed for reproducing the results, but rather reference the parts of
% the main paper, appendix, and supplemental materials that will help with
% reproducibility. For example, for novel models or algorithms, a link to an
% anonymous downloadable source code can be submitted as supplementary materials;
% for theoretical results, clear explanations of any assumptions and a complete
% proof of the claims can be included in the appendix; for any datasets used in
% the experiments, a complete description of the data processing steps can be
% provided in the supplementary materials. Each of the above are examples of
% things that can be referenced in the reproducibility statement.
Section~\ref{sec:problem-formulation} specifies the evolution procedure
and IDM updates.
Appendix~\ref{app:experimental-details} documents initialization,
instruction templates, model configurations, task splits, evaluation
protocols, prediction losses, and statistical reporting.
Appendices~\ref{app:run-level-results} and~\ref{app:evaluation-cost-analysis}
provide individual-run results and token-budget accounting, respectively.
Our code repository is at
\url{https://anonymous.4open.science/r/Predictive-self-evolution/}.

\bibliography{iclr2027_conference}
\bibliographystyle{iclr2027_conference}

\appendix

\section{Experimental Details}
\label{app:experimental-details}

Our implementation builds on the released HyperAgents
codebase \citep{zhang2026hyperagents}.
We retain its parent-selection rule and domain evaluators.

\subsection{Initialization and Instructions}
\label{app:initial-agents}
\label{app:idm-instructions}

Following HyperAgents \citep{zhang2026hyperagents},
we keep the initial programs simple to leave room
for strategies to develop through evolution.
The instructions specify each component's role,
available inputs and tools, and required outputs,
with separate meta-agent instructions distinguishing
Hyperagent modification from IDM updates.
We use this initialization approach across domains, adapting task descriptions, inputs, and required outputs to each domain.
The initial instruction templates are shown below.

\subsubsection{Initial Components}
\label{app:initial-components}

\paragraph{Task agent.}
For paper review, robotics, and mathematical grading, the initial task agent receives domain-specific inputs through the following instruction:

\begin{lstlisting}[style=idmprompt] 
You are an agent.

Task input:
{inputs}

Return one JSON object. The response value must be a string containing your answer.
Do not wrap the JSON in Markdown fences or XML tags.
{
    "response": "your answer"
}
\end{lstlisting}
The coding task agent additionally receives repository-editing instructions with access to shell and file-editing tools.

\paragraph{Meta agent.}
The meta agent receives the codebase location and access to previous evaluation results.
When available, archived predictions and observed outcomes are also accessible, and a prediction-error report is included in the request.
The initial instruction includes:

\begin{lstlisting}[style=idmprompt]
Modify any part of the codebase at `{repo_path}`. Evaluation results and prior generation outputs are stored at `{eval_path}`.

A read-only audited reaction-history substrate is available at `{reaction_archive_path}`; inspect it only when useful.

This is the task-improvement phase: the modification you produce now is what gets evaluated on the benchmark, so it must change how tasks are solved. `intervention_model.py` forecasts the evaluation state a modification produces and is revised in a separate phase; edits to it are discarded here.
\end{lstlisting}

For IDM updates, the same meta agent starts a fresh
session with the current IDM, archived modifications
and their predicted and observed outcomes, and a
prediction-error report when available.
Edits are restricted to the IDM.
The update instruction includes:

\begin{lstlisting}[style=idmprompt]
The benchmark modification for this generation has already been evaluated; edits you make now are not scored on the benchmark and take effect for later generations. Edits to any other file are discarded.
\end{lstlisting}

\paragraph{IDM.}
Across domains, the IDM is instructed to predict evaluation outcomes from the proposed modification and its expected behavioral consequences.
A representative initial system prompt is:

\begin{lstlisting}[style=idmprompt]
You are the IDM of a self-evolving agent.

Predict the evaluation outcomes of the proposed code modification.
Ground your predictions in the affected code and its expected
behavioral consequences. Include a brief explanation.
Do not write code.
\end{lstlisting}

\subsubsection{Revision Feedback to the Meta Agent}
\label{app:revision-feedback}

An additional revision pass continues the same meta-agent
session after a draft has been produced, preserving
the draft and interaction history.
Reflective HyperAgents requests this pass for every
draft without invoking the IDM, using the following
instruction:

\begin{lstlisting}[style=idmprompt]
Reconsider the draft modification using only the information already available. Check for likely failure modes and revise the draft if warranted. No new evaluation feedback is available.
\end{lstlisting}

When the revision rule requests an additional revision pass,
Selective IDM uses the same reconsideration instruction
and appends the IDM feedback:

\begin{lstlisting}[style=idmprompt]
Prospective IDM prediction (information unavailable to the reflection control; use it as a hypothesis, not as evaluation feedback):

--- BEGIN IDM PREDICTION ---
{delivered forecast}
--- END IDM PREDICTION ---
\end{lstlisting}

IDM Reflection uses the same forecast feedback for
every draft.
Prediction formats, losses, calibration, and update
validation are described in
Appendix~\ref{app:idm-implementation}.

\subsection{Domains and Evaluation Protocols}
\label{app:domains}
We use the domain evaluators of HyperAgents \citep{zhang2026hyperagents}, with task splits and evaluation scores summarized in Table~\ref{tab:domain-protocols}.
The task scores $s_\tau(h)$ in \eqref{eq:evaluation-score} are specified below.

\begin{table}[H]
    \centering
    \caption{
    Task splits and scores used in our experiments.
    Robotics counts refer to repeated reward-design trials
    on the same control task.
    }
    \label{tab:domain-protocols}
    \small
    \setlength{\tabcolsep}{4pt}
    \begin{tabular}{@{}lcccl@{}}
        \toprule
        Domain & Train & Validation & Held-out & Evolution score \\
        \midrule
        Polyglot coding
            & 60 & -- & 166 & Resolve rate \\
        Paper review
            & 100 & 100 & -- & Validation accuracy \\
        Robotics reward design
            & 6 & -- & -- & Training fitness \\
        Mathematical grading
            & 100 & 100 & 100 & Validation accuracy \\
        \bottomrule
    \end{tabular}
\end{table}

% \paragraph{Coding.}
% Polyglot \citep{gauthier2024polyglot} requires the task agent
% to produce a patch that passes the task's test suite.
% The task score $s_\tau(h)$ is 1 if the submitted patch passes
% the benchmark tests and 0 otherwise.
% We report pass@1 resolve rate, with no feedback from the
% final benchmark evaluation returned during the attempt.
% Evaluation starts with a fixed subset of 10 evolution tasks
% and continues on the remaining 50 if the resolve rate on
% that subset exceeds 40\%.
% Archive scores use all 60 tasks as the denominator,
% with unevaluated tasks contributing zero.
% The same preliminary subset is used across methods.
% After evolution, we select the evaluated descendant with the highest evolution score in each run, breaking ties by the earliest generation.
% The selected agent is evaluated on 166 disjoint held-out tasks, which are not used for modification or selection.
\paragraph{Coding.}
Polyglot \citep{gauthier2024polyglot} requires the task agent
to produce a patch that passes the task's test suite.
The task score $s_\tau(h)$ is 1 if the submitted patch passes
the benchmark tests and 0 otherwise.
We report pass@1 resolve rate, with no feedback from the
final benchmark evaluation returned during the attempt.
Evaluation starts with a fixed subset of 10 evolution tasks,
shared across methods, and proceeds to full evaluation
if at least four preliminary tasks are resolved.
Archive scores use all 60 tasks as the denominator,
with unevaluated tasks contributing zero.
For coding, both the predicted score and the observed
parent score in \eqref{eq:revision-gate} are computed
on the same ten preliminary tasks.
Forecast evaluation uses only tasks with observed outcomes,
matching predictions and observations on the same instances.
After evolution, we select the evaluated descendant with
the highest evolution score in each run, breaking ties
by the earliest generation.
The selected agent is evaluated on 166 disjoint held-out
tasks, which are not used for modification or selection.

\paragraph{Joint paper review and robotics.}
Paper review and robotics are optimized within the same
evolution run.
For paper review, the task agent receives a research paper
and predicts acceptance or rejection \citep{zhao2026apres}.
The task score $s_\tau(h)$ is 1 if the prediction matches
the recorded conference decision and 0 otherwise.

For robotics reward design, the task agent generates a Python reward function for quadruped velocity tracking in Genesis \citep{genesis2024}.
Each trial trains a new control policy using proximal policy optimization.
The task score $s_\tau(h)$ is the trial fitness, which aggregates the fixed tracking score $\exp[-4(v_x-v_x^*)^2]$, where $v_x$ and $v_x^*$ are the achieved and commanded forward velocities, following the upstream evaluation procedure.
The robotics score averages these task scores, with task-agent failures scored as zero and identified infrastructure failures excluded.

Each Hyperagent's joint score averages its paper-review
validation accuracy and robotics training fitness
with equal weight.
Preliminary evaluation uses 10 training and 10 validation
papers and three robotics trials.
If both domains obtain positive preliminary training
scores, evaluation expands to 100 training papers,
100 validation papers, and six robotics trials.
Otherwise, the preliminary paper-review validation accuracy
and mean robotics fitness are multiplied by $10/100$
and $3/6$, respectively, before computing the joint score.
% \paragraph{Paper review.}
% The task agent receives a research paper and predicts
% acceptance or rejection \citep{zhao2026apres}.
% The task score $s_\tau(h)$ is 1 if the prediction matches
% the recorded conference decision and 0 otherwise.
% The preliminary evaluation uses 10 training papers and
% is passed if at least one decision is correct.
% Full training evaluation covers all 100 training papers.
% Accuracy on the separate 100-paper validation split
% contributes to the joint objective.

% \paragraph{Robotics reward design.}
% The task agent generates a Python reward function for a
% quadruped velocity-tracking task in Genesis \citep{genesis2024}.
% Each reward-design trial generates a reward function and
% trains a new control policy using proximal policy optimization.
% Trial fitness aggregates the fixed tracking score
% $\exp[-4(v_x-v_x^*)^2]$, where $v_x$ and $v_x^*$
% are the achieved and commanded forward velocities,
% following the upstream evaluation procedure.
% The preliminary evaluation uses three trials and is passed
% if at least one achieves nonzero fitness.
% Full evaluation adds three further trials.
% The task score $s_\tau(h)$ averages fitness over all six
% trial slots, with failed or unexecuted trials contributing zero.

\paragraph{Mathematical grading.}
IMO-GradingBench \citep{luong2025towards} provides an Olympiad-level problem, a proposed solution, reference solutions, and grading guidelines.
The task agent assigns a grade from $\{0,1,6,7\}$, corresponding to incorrect, partially correct, almost correct, and correct solutions, respectively.
The task score $s_\tau(h)$ is 1 if the predicted grade exactly matches the expert grade and 0 otherwise.
Preliminary evaluation uses 10 training and 10 validation tasks and expands to 100 tasks in each split if at least one training prediction matches the expert grade.
Validation accuracy guides evolution.
For held-out evaluation, we select the descendant with the highest validation accuracy among those evaluated on all 100 validation tasks in each run, breaking ties by the earliest generation.

\subsection{Experimental Conditions}
\label{app:experimental-conditions}

Each experimental condition consists of five runs of 40 generations.
Within each setting, methods use the same initial task and meta agents, task splits, evaluators, and parent-selection rule.
They also share the same model assignments and initial model-call configurations.
The underlying LLM weights remain fixed throughout evolution.
For conditions with an evolving IDM, update proposals are scheduled every $K=5$ generations.

\paragraph{Parent selection.}
We use the parent-selection rule of HyperAgents
\citep{zhang2026hyperagents}.
The initial Hyperagent and valid evaluated descendants
are eligible for selection.
At generation $t$, let $\bar y_t$ denote the mean of the
three highest evaluation scores among eligible parents,
using all available scores when fewer than three
are eligible.
For an archived Hyperagent $h$, let $n_t(h)$ denote
the number of its direct children recorded in the archive.
We sample the parent with probability proportional to
\begin{equation}
    w_t(h)
    =
    \exp\!\left[-\left(\frac{n_t(h)}{8}\right)^3\right]
    \frac{1}{1+\exp\!\left[-10\left(y(h)-\bar y_t\right)\right]},
    \label{eq:parent-selection-weight}
\end{equation}
where $y(h)$ is the aggregate evaluation score defined in
\eqref{eq:evaluation-score}.
The weights are normalized over eligible parents.
This rule favors higher-scoring Hyperagents while
reducing repeated expansion of parents that have
already produced many children.

\paragraph{Models and generation settings.}
Table~\ref{tab:llm-configuration} summarizes the initial model assignments.
GPT-5.4 calls use high reasoning effort.
Across domains, the shared LLM interface defaults to an output limit of 16{,}384 tokens per call.
Temperature is set to 0.0 for Claude and left unspecified for GPT-4o and o3-mini.
The shared tool framework defaults to a maximum of 40 tool calls per attempt.

\begin{table}[H]
    \centering
    \caption{Initial LLM assignments across experimental settings.}
    \label{tab:llm-configuration}
    \small
    \begin{tabular}{@{}lll@{}}
        \toprule
        Setting & Task agent & Meta agent and IDM \\
        \midrule
        Polyglot coding
            & o3-mini & Claude Sonnet 4.6 \\
        Joint: paper review
            & GPT-4o & GPT-5.4 \\
        Joint: robotics
            & GPT-5.4 & GPT-5.4 \\
        Mathematical grading
            & GPT-4o & GPT-5.4 \\
        \bottomrule
    \end{tabular}
\end{table}

\paragraph{Baselines and variants.}
On Polyglot, we compare Selective IDM with HyperAgents, Reflective HyperAgents, and two IDM variants, as summarized in Table~\ref{tab:variant-configuration}.
HyperAgents proceeds from a proposed modification to evaluation without an additional revision pass.
Reflective HyperAgents and IDM Reflection both revise every draft once using the same reconsideration instruction, with IDM Reflection additionally receiving the IDM forecast.
Frozen Selective IDM fixes the forecasting program while retaining selective revision and online bias calibration.
On Polyglot, all IDMs share the same initialization.

\begin{table}[H]
    \centering
    \caption{
    Baselines and variants compared on Polyglot.
    ``Always'' requests one revision for every draft.
    ``Selective'' requests one revision according to
    the revision rule.
    }
    \label{tab:variant-configuration}
    \small
    \setlength{\tabcolsep}{4pt}
    \begin{tabular}{@{}lccc@{}}
        \toprule
        Method & IDM feedback & Revision rule & IDM updates \\
        \midrule
        HyperAgents
            & No & None & -- \\
        Reflective HyperAgents
            & No & Always & -- \\
        IDM Reflection
            & Yes & Always & Yes \\
        Frozen Selective IDM
            & Yes & Selective & No \\
        Selective IDM
            & Yes & Selective & Yes \\
        \bottomrule
    \end{tabular}
\end{table}

Joint evolution compares HyperAgents with Selective IDM.
Mathematical grading compares HyperAgents, Selective IDM,
and Transferred IDM, a variant of Selective IDM initialized
with an IDM evolved jointly on paper review and robotics.
Only the IDM is transferred.
Both conditions use the same initial task and meta
agents, start with fresh calibration state and interaction
history, and continue updating the IDM.

We selected an evolved IDM based on source-domain
prediction improvements and a preference for fewer
domain-specific heuristics.
This selection preceded the mathematical-grading
experiments, and the same IDM initializes all five
transfer runs.
Its evolved forecasting instructions include:

\begin{lstlisting}[style=idmprompt]
Start from the parent's recorded state or metric for each
instance, but treat that as a baseline to update rather
than a label to copy.
...
Do not anchor on parent labels or opaque instance IDs when
the same changed code governs many instances together.
\end{lstlisting}

These instructions are transferred unchanged, without
adding mathematical-grading-specific heuristics.
Tool-budget finalization and final-answer handling are
aligned with those of the initial IDM.

\subsection{IDM Implementation and Evaluation}
\label{app:idm-implementation}

The IDM predicts task outcomes using domain-specific
formats and losses.
Its forecasting program is editable, while the outcome
definitions and prediction losses remain fixed.

\subsubsection{Prediction Formats and Losses}
\label{app:idm-prediction-formats}

\paragraph{Prediction outputs.}
For coding, paper review, and mathematical grading,
the IDM returns a normalized probability distribution
over the outcome classes in
Table~\ref{tab:idm-categorical-outcomes}
for each evaluation task.
The predicted task score is the probability of
\texttt{resolved} for coding and \texttt{correct}
for paper review and mathematical grading.

\begin{table}[H]
    \centering
    \caption{Categorical outcomes predicted by the IDM.}
    \label{tab:idm-categorical-outcomes}
    \small
    \setlength{\tabcolsep}{5pt}
    \ \renewcommand{\arraystretch}{1.15}
    \begin{tabular}{@{}llp{0.49\linewidth}@{}}
        \toprule
        Domain & Outcome & Meaning \\
        \midrule
        Coding
            & \texttt{resolved}
            & A patch is produced and passes the benchmark tests. \\
        & \texttt{unresolved}
            & A patch is produced but does not pass the benchmark tests. \\
        & \texttt{incomplete}
            & The task-agent run does not finish. \\
        & \texttt{empty\_patch}
            & The task-agent run finishes without producing a patch. \\
        \midrule
        \shortstack[l]{Paper review and\\mathematical grading}
            & \texttt{correct}
            & The task agent predicts the correct decision or grade. \\
        & \texttt{incorrect}
            & The task agent returns a valid but incorrect decision or grade. \\
        & \texttt{invalid}
            & The task-agent response is missing or unusable. \\
        \bottomrule
    \end{tabular}
\end{table}

For each robotics trial, the IDM predicts the probability
$p_{\mathrm{invalid}}$ that the generated reward function
is unusable and the expected fitness $m$ conditional on
the reward function being usable, with both values in $[0,1]$.
Since an unusable reward function receives zero fitness,
the predicted trial score is $(1-p_{\mathrm{invalid}})m$.

In paper review, robotics, and mathematical grading,
the IDM is instructed to provide brief structured
explanations alongside its forecasts, while explanations
are optional in coding.
Only the numerical forecasts are scored.

\paragraph{Task-level prediction losses.}
For categorical predictions, let $\mathcal C_d$ denote
the outcome classes in domain $d$,
$\mathbf p=(p_c)_{c\in\mathcal C_d}$ the predicted
probability distribution, and $c^*$ the observed outcome.
The multiclass Brier loss \citep{brier1950verification} is
\begin{equation}
    \ell_{\mathrm{Brier}}(\mathbf p,c^*)
    \doteq
    \sum_{c\in\mathcal C_d}
    \left(p_c-\mathbf 1\{c=c^*\}\right)^2.
    \label{eq:appendix-categorical-loss}
\end{equation}
Coding uses this loss directly.
Paper review and mathematical grading use half of
this loss, placing it in $[0,1]$ and matching the range
of the robotics loss below.

For robotics, let $s$ denote the observed trial fitness
and let $z_{\mathrm{invalid}}$ be 1 if the reward function
is unusable and 0 otherwise.
The prediction loss is
\begin{equation}
    \ell_{\mathrm{robotics}}
    \doteq
    \frac12
    \left[
        \left(p_{\mathrm{invalid}}-z_{\mathrm{invalid}}\right)^2
        +
        \left((1-p_{\mathrm{invalid}})m-s\right)^2
    \right].
    \label{eq:appendix-robotics-loss}
\end{equation}
The two terms measure error in the invalidity probability
and predicted trial fitness, respectively.

\paragraph{Aggregate prediction loss.}
For domain $d$, let $N_d$ denote the number of scored
predictions across the evaluation block.
For IDM $I$, let $\ell_{d,i}(I)$ denote the loss of
prediction $i$, using the corresponding domain loss
defined above.
The mean prediction loss within the domain is
\begin{equation}
    \mathcal L_d(I)
    \doteq
    \frac{1}{N_d}\sum_{i=1}^{N_d}\ell_{d,i}(I).
\end{equation}
For coding and mathematical grading,
$\mathcal L(I)=\mathcal L_d(I)$.
For joint evolution, we give the two domains equal weight:
\begin{equation}
    \mathcal L(I)
    \doteq
    \frac12\left[
        \mathcal L_{\mathrm{paper}}(I)
        +
        \mathcal L_{\mathrm{robotics}}(I)
    \right].
    \label{eq:appendix-aggregate-prediction-loss}
\end{equation}

\subsubsection{Update Validation and Calibration}
\label{app:idm-update-validation}

Following Section~\ref{sec:evolving-idm}, we compare
candidate and current IDMs on the same $K$ records,
collected after candidate construction while the current
IDM remains deployed.
Let $\mathcal V_k$ index the validation records for update $k$,
and let $\mathcal S_j$ denote the observed evaluation
instances in record $j$.
We write $y(h;\mathcal S_j)$ and
$\widehat y_j(I,h;\mathcal S_j)$ for observed and predicted
scores computed on this set, with equal domain weights
in joint evolution.
For either IDM $I$, we compute score prediction errors
before applying the calibration offset, and define
the block's mean bias and centered mean squared error as
\begin{equation}
\begin{aligned}
    e_j(I)
    &\doteq
    \widehat y_j\!\left(
        I,h_{p_j}^{\delta_j};\mathcal S_j
    \right)
    - y(h_{j+1};\mathcal S_j), \\
    b_k(I)
    &\doteq
    \frac{1}{K}\sum_{j\in\mathcal V_k} e_j(I), \\
    \operatorname{CMSE}_k(I)
    &\doteq
    \frac{1}{K}\sum_{j\in\mathcal V_k}
    \left(e_j(I)-b_k(I)\right)^2.
\end{aligned}
\label{eq:appendix-cmse}
\end{equation}
CMSE measures the residual score prediction error after
removing the mean bias within the block
\citep{taylor2001summarizing}.
The candidate replaces the current IDM only if it
lowers the aggregate prediction loss $\mathcal L(I)$
without increasing CMSE.
The comparison requires complete, valid forecasts from
both IDMs for all required instances.
Outcome probabilities must form normalized distributions,
and numerical predictions must be finite.
Figure~\ref{fig:idm-improvement}(a) includes all 28 comparisons
with complete, valid forecasts from both IDMs within the
five 40-generation Polyglot runs.
Six archived comparisons from post-budget continuations
are excluded.

The calibration offset is initialized to $b=0$ at the
start of a run.
Before each revision decision, $b$ is re-estimated
as the mean signed score prediction error over valid
past records from the current IDM.
Errors are computed from uncalibrated predicted scores,
with positive values indicating overprediction.
For coding, this online estimate uses predictions and
observed outcomes on the same fixed ten-task set as
the revision decision, even when additional tasks
are evaluated.
Upon promotion, the candidate's validation-block bias
$b_k(I)$ initializes $b$ until valid online records
from the new IDM become available.
Frozen Selective IDM follows the same online calibration
rule while keeping its forecasting program fixed.
% The revision rule's calibration offset is initialized
% to $b=0$.
% Upon promotion, the candidate's mean error on the
% validation block becomes the new offset $b$.
% This offset remains fixed until another update is promoted.

\subsection{Budgets, Metrics, and Statistical Reporting}
\label{app:budgets-statistics}

We compute task-performance and prediction-error metrics
within each evolution run before summarizing results
across runs.

\paragraph{AUBC.}
Generation-based AUBC follows \eqref{eq:aubc}.
Under both generation and token budgets, the performance
frontier excludes the initial Hyperagent and starts at zero.
It changes only after a descendant's evaluation completes
and remains constant between evaluations.

Token budgets count input and output tokens used for
agent evolution and task evaluation, including IDM
forecasting, updates, and validation.
For each comparison, the common budget is the median
total token cost of five complete 40-generation runs
of the reference baseline.
We use HyperAgents as the reference for the main
comparisons and Reflective HyperAgents for the
coding ablations.
This anchors the comparison to the typical cost of
a complete baseline run while reducing sensitivity
to unusually high or low run costs.

Let $B$ denote the common token budget and $F_r(c)$
the best-so-far score in run $r$ at cumulative token
use $c$.
We compute
\begin{equation}
    \operatorname{AUBC}_{\mathrm{token},r}(B)
    \doteq
    \frac{1}{B}\int_0^B F_r(c)\,\mathrm{d}c.
    \label{eq:token-aubc}
\end{equation}
If a run finishes below the token budget, its final
frontier value is held constant up to the budget.
Budget values and additional matched-token comparisons are provided in Appendix~\ref{app:evaluation-cost-analysis}.

\paragraph{Statistical reporting.}
We report medians over five complete runs.
Where shown, 95\% bootstrap confidence intervals
are obtained by resampling runs with replacement.
For curves, entire runs are resampled together across
budget points, and the intervals are pointwise.
% We use 1{,}000 resamples for coding task-performance
% plots and 1{,}000 for mathematical-grading
% task-performance plots.
Coding and mathematical-grading task-performance plots
use 1{,}000 resamples.
% For the differences of medians in
% Table~\ref{tab:ablation-differences}, we enumerate all
% $5^5$ ordered run resamples per method and combine
% the two methods' resamples independently.
% The 2.5th and 97.5th percentiles of this exact bootstrap
% distribution define each marginal interval.
% Runs are not paired across methods.
Intervals in Table~\ref{tab:ablation-differences} use an exact
percentile bootstrap with runs resampled independently
within each method.

\section{Additional Results and Analysis}
\label{app:additional-results}

\subsection{Quality of Generated Descendants}
\label{app:descendant-quality}

For each coding run, we compute every descendant's resolve rate
on the same ten preliminary tasks and average these rates over
all 40 generations, including generations that solve no tasks.
We also measure the fraction of generations that reach full
evaluation by solving at least four of the ten tasks.
Table~\ref{tab:descendant-quality} summarizes all five configurations.

\begin{table}[t]
    \centering
    \caption{
    \textbf{Quality of generated coding descendants.}
    Values are percentages, reported as medians over five runs.
    Each metric is computed within a run over all 40 generations.
    }
    \label{tab:descendant-quality}
    \small
    \setlength{\tabcolsep}{4pt}
    \begin{tabular}{@{}lrr@{}}
        \toprule
        Method & \shortstack{Mean resolve\\rate (\%)}
            & \shortstack{Full-evaluation\\generations (\%)} \\
        \midrule
        HyperAgents & 17.00 & 5.0 \\
        Reflective HyperAgents & 29.25 & 35.0 \\
        IDM Reflection & 36.50 & 57.5 \\
        Frozen Selective IDM & 28.25 & 30.0 \\
        Selective IDM & 29.50 & 45.0 \\
        \bottomrule
    \end{tabular}
\end{table}

% Compared with HyperAgents, Selective IDM increases the median
% mean resolve rate from 17.00\% to 29.50\% and reduces the median
% fraction of zero-solve generations from 27.5\% to 7.5\%.
% The improvement therefore extends beyond the best agents
% captured by AUBC to the descendants generated throughout evolution.

% When both methods revise every draft once, IDM Reflection
% achieves a higher median mean resolve rate than Reflective
% HyperAgents (36.50\% versus 29.25\%) and a higher median fraction
% of generations reaching full evaluation (57.5\% versus 35.0\%).
% This comparison supports the value of predictive feedback
% beyond an additional revision alone.
% Among the selective variants, Selective IDM also reaches full
% evaluation in a higher median fraction of generations than
% Frozen Selective IDM (45.0\% versus 30.0\%).

Selective IDM improves the quality of descendants produced
throughout coding evolution.
The median run-level mean resolve rate increases from 17.00\%
for HyperAgents to 29.50\% for Selective IDM, while the median
fraction of generations reaching full evaluation rises from
5.0\% to 45.0\%.
These gains show that predictive guidance improves not only
the best agents tracked by AUBC, but also the broader set of
descendants generated during evolution.

Predictive feedback also improves descendant quality when both
methods revise every draft once.
IDM Reflection achieves a median run-level mean resolve rate
of 36.50\%, compared with 29.25\% for Reflective HyperAgents,
and more generations reach full evaluation.
Selective IDM likewise exceeds Frozen Selective IDM on both
measures, supporting the contribution of IDM updates to the
quality of generated descendants.

\subsection{Token Costs and Matched-Token Ablations}
\label{app:evaluation-cost-analysis}

Following HyperAgents \citep{zhang2026hyperagents},
we report token costs for agent evolution and task evaluation,
% estimated retrospectively from interaction logs.
estimated from interaction logs.
Table~\ref{tab:token-costs} summarizes mean costs over
five complete 40-generation runs.
Using HyperAgents as the reference gives matched-token
budgets of 44.21M for coding, 111.98M for joint evolution,
and 69.36M for mathematical grading.

\begin{table}[t]
    \centering
    \caption{
    \textbf{Token costs of complete evolution runs.}
    Values are mean estimated token counts over
    five 40-generation runs, in millions.
    Initialization, independent held-out evaluation,
    and post-hoc analyses are excluded.
    }
    \label{tab:token-costs}
    \small
    \setlength{\tabcolsep}{4pt}
    \begin{tabular}{@{}llr@{}}
        \toprule
        Setting & Method & Mean tokens (M) \\
        \midrule
        Coding & HyperAgents & 49.45 \\
        & Reflective HyperAgents & 138.56 \\
        & Frozen Selective IDM & 134.62 \\
        & IDM Reflection & 207.58 \\
        & Selective IDM & 170.22 \\
        \midrule
        Joint & HyperAgents & 101.67 \\
        & Selective IDM & 140.61 \\
        \midrule
        Math & HyperAgents & 69.46 \\
        & Selective IDM & 99.13 \\
        & Transferred IDM & 104.15 \\
        \bottomrule
    \end{tabular}
\end{table}

\begin{figure}[t]
    \centering
    \includegraphics[width=\linewidth]
        {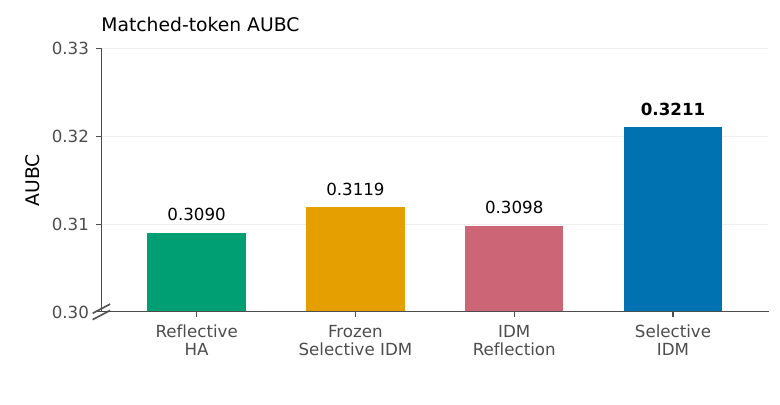}
    \caption{
    \textbf{Selective IDM retains its advantage
    under a matched token budget.}
    Bars show median AUBC over five runs on Polyglot
    at a common budget of 145.79M tokens.
    The vertical axis is truncated.
    }
    \label{fig:coding-token-ablations}
\end{figure}

For the coding ablations, Reflective HyperAgents serves
as the reference, giving all four configurations a common
budget of 145.79M tokens.
Selective IDM achieves the highest median AUBC under
this budget, exceeding Reflective HyperAgents,
Frozen Selective IDM, and IDM Reflection by
3.9\%, 2.9\%, and 3.6\%, respectively
(Figure~\ref{fig:coding-token-ablations}).
Its mean cost over complete 40-generation runs is also
18.0\% lower than that of IDM Reflection.
These results support using an evolving IDM to
allocate revision effort selectively, improving
performance under a matched budget while reducing
the cost relative to unconditional IDM-guided revision.

\subsection{Run-Level Results}
\label{app:run-level-results}

\begin{figure}[t]
    \centering
    \includegraphics[width=\linewidth]{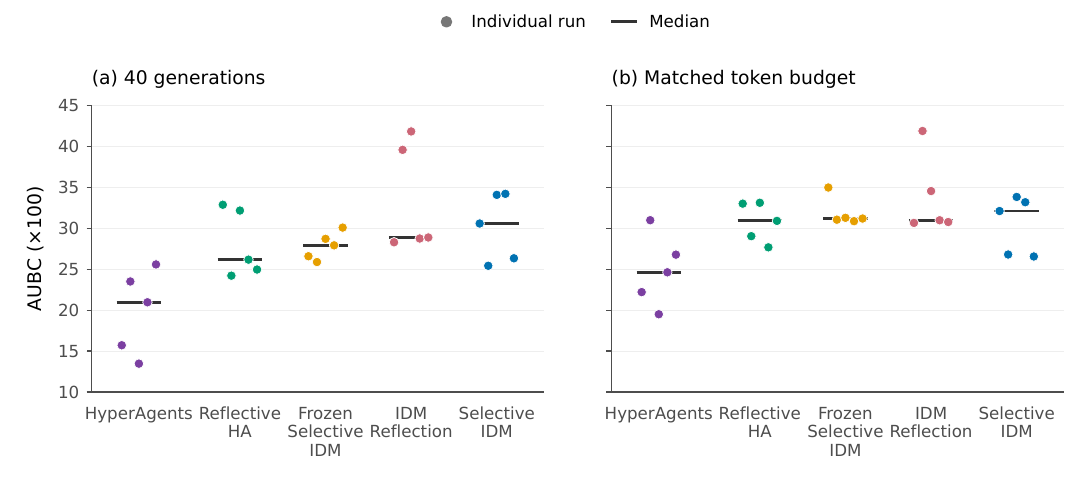}
    \caption{
    \textbf{Individual-run results for coding.}
    AUBC on Polyglot under (a) 40 generations and (b) a common
    budget of 145.79M estimated tokens.
    The token budget is the median cost of five complete
    Reflective HyperAgents runs and is shared by all five methods.
    AUBC values are multiplied by 100.
    The vertical axes are truncated.
    }
    \label{fig:run-level-coding}
\end{figure}

\begin{figure}[t]
    \centering
    \includegraphics[width=\linewidth]{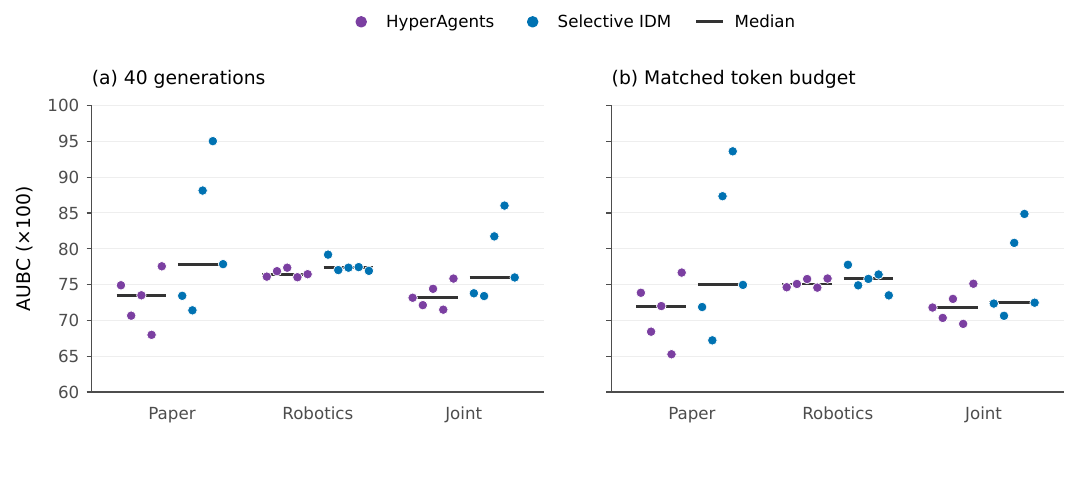}
    \caption{
    \textbf{Individual-run results for joint evolution.}
    AUBC for paper review, robotics, and the joint objective
    under (a) 40 generations and (b) a common budget of
    111.98M estimated tokens.
    AUBC values are multiplied by 100.
    The vertical axes are truncated.
    }
    \label{fig:run-level-joint}
\end{figure}

\begin{figure}[t]
    \centering
    \includegraphics[width=\linewidth]{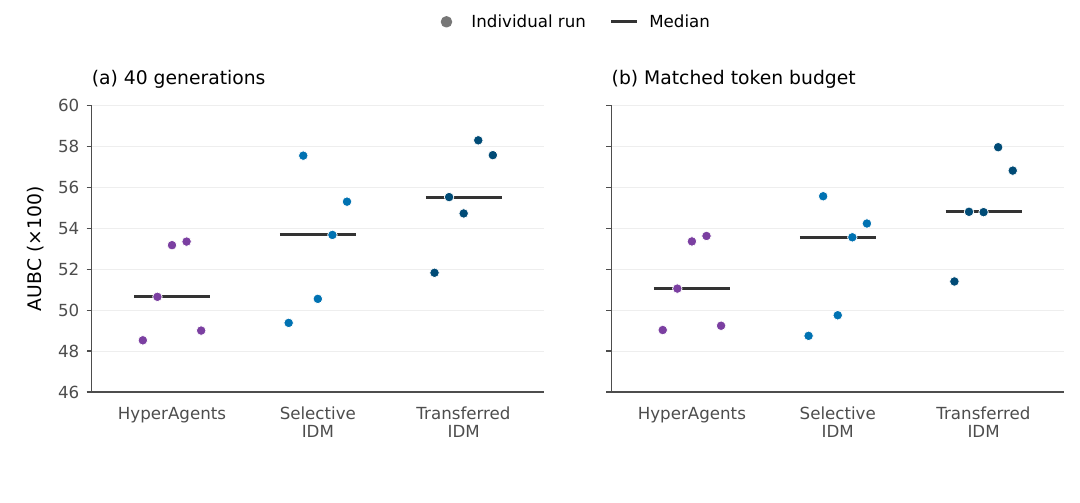}
    \caption{
    \textbf{Individual-run results for mathematical grading.}
    AUBC under (a) 40 generations and (b) a common budget
    of 69.36M estimated tokens.
    AUBC values are multiplied by 100.
    The vertical axes are truncated.
    }
    \label{fig:run-level-math}
\end{figure}

The individual-run results complement the aggregate comparisons
in Section~\ref{sec:experiments}.
Figures~\ref{fig:run-level-coding}, \ref{fig:run-level-joint},
and~\ref{fig:run-level-math} report AUBC under 40-generation
and matched-token budgets.
Each method contributes five complete runs; circles show
individual results and horizontal segments mark their median.

\paragraph{Coding.}
Selective IDM's advantage over HyperAgents extends across
the five runs.
All five Selective IDM runs exceed the HyperAgents median
under both budgets (Figure~\ref{fig:run-level-coding}).
Selective IDM also achieves the highest median AUBC among
the revision strategies under both budgets.
Table~\ref{tab:ablation-differences} reports the differences
in median AUBC relative to Reflective HyperAgents,
Frozen Selective IDM, and IDM Reflection.

% \paragraph{Uncertainty in coding ablations.}
% Table~\ref{tab:ablation-differences} reports the difference
% between Selective IDM's median AUBC and each ablation's
% median under both budgets.
% All six point estimates favor Selective IDM; the intervals
% include zero, reflecting uncertainty across five runs
% per method.
% \paragraph{Coding ablation comparisons.}
% Table~\ref{tab:ablation-differences} summarizes differences
% between Selective IDM and the alternative configurations
% under both budgets.

\begin{table}[t]
    \centering
    \caption{
    \textbf{Median coding AUBC differences: Selective IDM minus each comparator.}
    Each entry is Selective IDM minus the comparator,
    followed by its 95\% run-level bootstrap interval.
    Values are absolute AUBC differences multiplied by 100,
    not relative percentage gains.
    The token budget is 145.79M for all methods.
    }
    \label{tab:ablation-differences}
    \small
    \setlength{\tabcolsep}{5pt}
    \begin{tabular}{@{}lrr@{}}
        \toprule
        Comparator & 40 generations & Matched tokens \\
        \midrule
        Reflective HyperAgents
            & $4.42\;[-6.54,\,9.25]$ & $1.21\;[-6.32,\,4.79]$ \\
        Frozen Selective IDM
            & $2.67\;[-3.29,\,7.62]$ & $0.92\;[-4.73,\,2.64]$ \\
        IDM Reflection
            & $1.71\;[-14.17,\,5.46]$ & $1.13\;[-9.78,\,2.85]$ \\
        \bottomrule
    \end{tabular}
\end{table}

\paragraph{Joint evolution.}
Selective IDM achieves higher median AUBC for paper review,
robotics, and the joint objective under both budgets
(Figure~\ref{fig:run-level-joint}).
The gains are observed in both constituent domains as well
as in the joint score used to guide evolution.

\paragraph{Mathematical grading.}
Transferred IDM achieves the highest median AUBC under both
budgets (Figure~\ref{fig:run-level-math}).
Under each budget, four of the five transfer runs exceed
the median of Selective IDM initialized from scratch.
The benefit of predictive experience is therefore observed
across multiple target-domain runs initialized with the
same source-domain IDM.

\subsection{Contribution of Modification Information}
\label{app:modification-information}

The IDM forecasts the consequences of a proposed change
using evidence from the modification itself.
To examine the value of this information, we compare
normal forecasts with a \emph{No-diff} condition that
withholds the diff and modified source code while
retaining parent code, parent outcomes, and historical
records.
The comparison covers generations 10, 20, and 30 from each
of five joint evolution runs, yielding 15 modifications,
with the same IDM version, model configuration, and recorded
outcomes used for each paired comparison.

\begin{figure}[t]
    \centering
    \includegraphics[width=\linewidth]
        {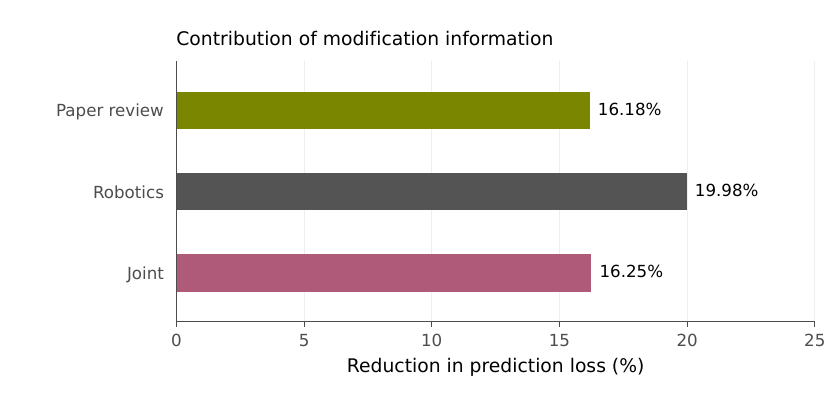}
    \caption{
    \textbf{Modification information improves IDM forecasts.}
    Relative reduction in mean prediction loss compared
    with No-diff, which withholds the diff and modified
    source code while retaining parent code, parent
    outcomes, and historical records.
    Comparisons use the same deployed IDM for each of
    15 evaluated modifications from five joint evolution runs.
    Joint loss weights the two domains equally before
    computing the relative reduction.
    }
    \label{fig:modification-information}
\end{figure}

Access to modification information reduces mean
prediction loss by 16.18\% for paper review and
19.98\% for robotics, yielding a 16.25\% reduction
in joint loss
(Figure~\ref{fig:modification-information}).
These gains support grounding forecasts in the proposed
code change, which provides information about its
consequences beyond the parent's observed performance
and historical outcomes.

% \subsection{Quality of Generated Descendants}
% \label{app:descendant-quality}

\subsection{Evolution Case Studies}
\label{app:evolution-case-studies}

\begin{figure}[t]
    \centering
    \includegraphics[width=\linewidth]
        {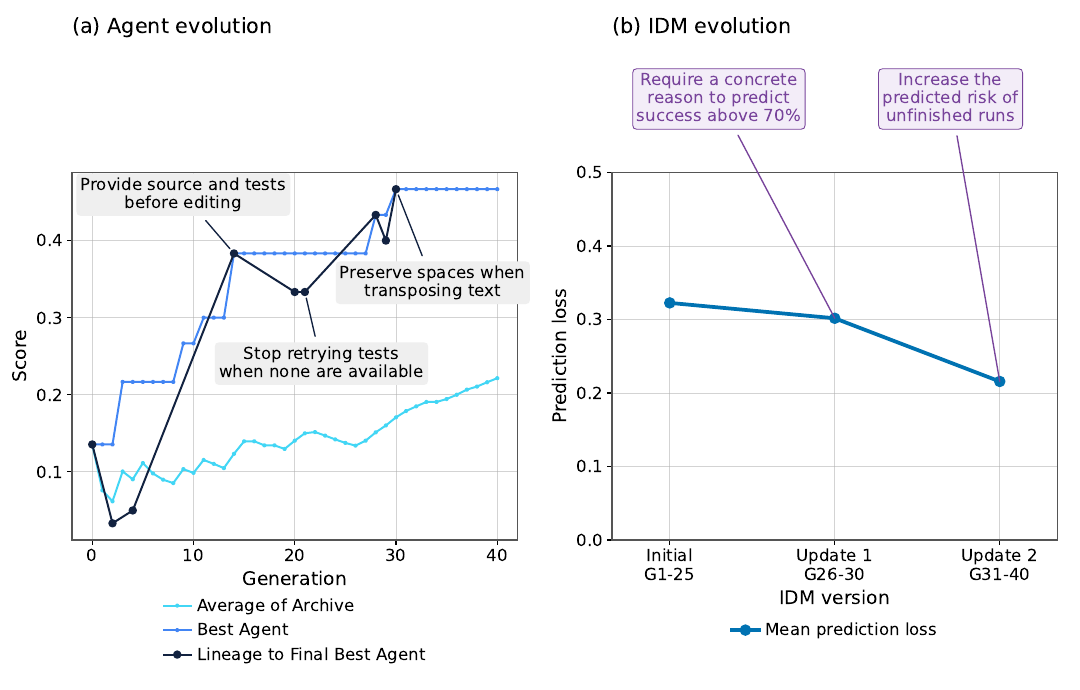}
    \caption{
    \textbf{Task strategies and forecasting rules evolve
    within the same run.}
    (a) Best and mean archive scores, together with
    the lineage leading to the final best agent.
    (b) Mean online prediction loss during each
    IDM version's deployment period.
    Annotations summarize selected changes to the
    task agent and the IDM.
    }
    \label{fig:evolution-path}
\end{figure}

Figure~\ref{fig:evolution-path} illustrates how task-solving
strategies and forecasting rules evolve within a coding run.
Along the lineage leading to the final best agent,
modifications provide source code and tests before editing,
avoid repeated test attempts when no tests are available,
and preserve spaces in text-transposition tasks.
These changes address both the agent's general workflow
and the requirements of specific tasks.

The IDM also develops more explicit criteria for forecasting
modification outcomes.
Its first update requires a concrete justification for
assigning success probabilities above 70\%, while the
second increases the predicted risk of unfinished runs.
Mean prediction loss decreases across the deployed versions.
This example illustrates how evolution can refine both
task-solving behavior and the forecasting rules used
to guide subsequent modifications.

\begin{figure*}[t]
    \centering
    \includegraphics[width=\textwidth]{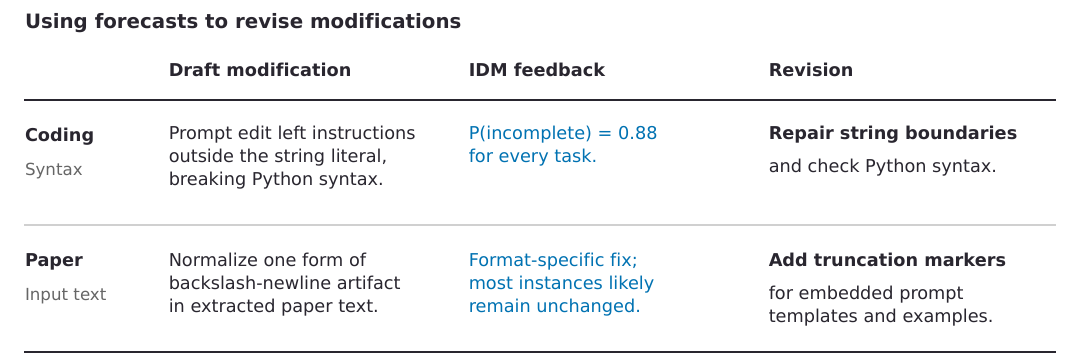}
    \caption{
    \textbf{Forecast-guided revision across tasks.}
    In Coding, a forecast of widespread incomplete execution
    prompts syntax repair.
    In Paper review, a forecast of limited applicability
    is followed by an expanded preprocessing change.
    }
    \label{fig:forecast-guided-revision}
\end{figure*}

\paragraph{Forecast-guided revision.}
Figure~\ref{fig:forecast-guided-revision} illustrates how
predictive feedback guides different kinds of revision.
In Coding, a prompt edit left instruction text outside
a string literal, introducing a syntax error.
The meta agent had noticed the damage but reached its
tool-call limit before completing the repair.
The IDM assigned a probability of 0.88 to
\texttt{incomplete} across the task set, and its
forecast triggered revision.
The meta agent responded by checking the file's syntax
and repairing the string boundaries.
The final version passed the syntax check and resolved
11 tasks.

In an earlier joint-evolution run, a Paper review
modification normalized one form of backslash-newline
artifact in extracted paper text.
The IDM noted that this format-specific change would
likely leave most instances unaffected.
During revision, the meta agent inspected saved paper
inputs and found embedded prompt-template text that
remained after the initial cleanup.
It extended the truncation markers to remove these
templates and examples.
Together, the cases illustrate how forecasts can guide
both execution repair and the scope of a proposed modification.

\begin{figure*}[t]
    \centering
    \includegraphics[width=\textwidth]{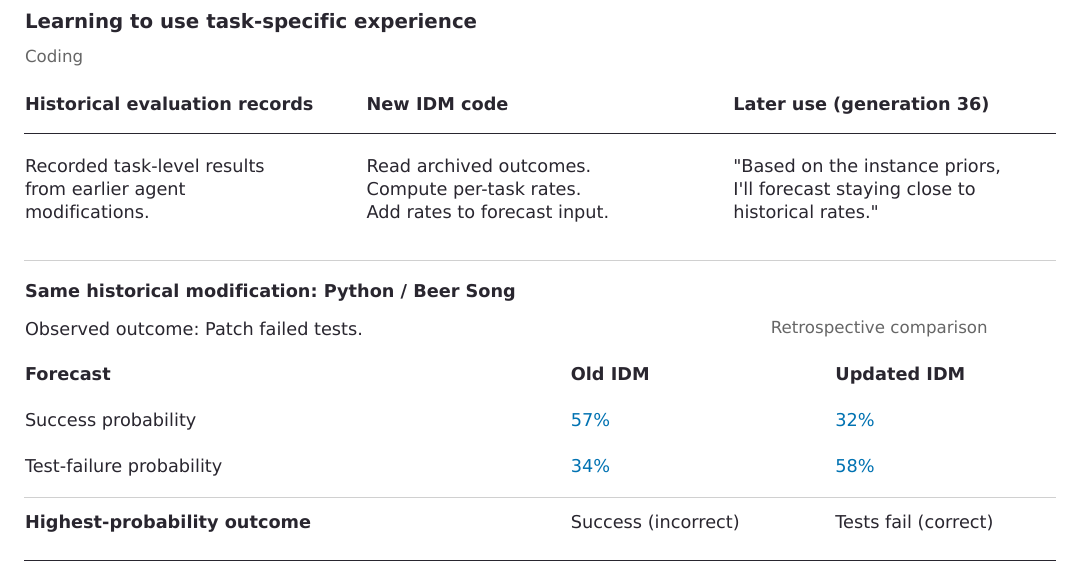}
    \caption{
    \textbf{Learning to use historical outcomes in forecasting.}
    The evolved IDM adds per-task outcome summaries to its
    forecast inputs; the quotation shows their use in a
    later online forecast.
    The lower panel separately compares the original and
    updated IDMs on the same historical modification.
    The updated IDM's highest-probability prediction matches
    the observed test failure.
    }
    \label{fig:idm-learning-case}
\end{figure*}

\paragraph{Learning to use historical experience.}
IDM evolution can also change how past outcomes inform
future predictions.
In the Coding case in Figure~\ref{fig:idm-learning-case},
the evolved program adds code that reads archived task
outcomes, computes per-task outcome frequencies, and
includes these summaries in subsequent forecast inputs.
Its updated instructions use these frequencies as priors
while allowing specific evidence from the proposed
modification to change the prediction.
A later deployed forecast explicitly invokes these priors.

The figure also compares the original and updated IDMs
on the same historical modification for Python Beer Song.
% In this retrospective comparison, the success probability
In this comparison, the success probability
decreases from 57\% to 32\%, while the probability of a
patch failing its tests increases from 34\% to 58\%.
The submitted patch failed its tests, so the
highest-probability prediction changes from incorrect
to correct.
On a separate validation block collected after the update
was proposed, the updated IDM also reduces Brier loss
from 0.6169 to 0.5951.
This case illustrates how evolution can encode historical
experience in the forecasting program and improve its
assessment of modification outcomes.

\end{document}